\documentclass{article}

    \PassOptionsToPackage{numbers, compress}{natbib}
\usepackage[preprint]{neurips_2026}

\usepackage[utf8]{inputenc} 
\usepackage[T1]{fontenc}    

\usepackage{comment}

\usepackage{amsmath,amssymb,mathtools}  
\usepackage{graphicx}                   
\usepackage{algorithm}                  
\usepackage{algorithmic}                
\usepackage{enumitem}   
\usepackage{subcaption}

\usepackage{soul}

\usepackage{booktabs} 
\usepackage{tabularx} 
\usepackage{multirow} 
\usepackage{pifont}   

\usepackage[table]{xcolor}
\definecolor{bestcell}{HTML}{EAF3FF}
\newcommand{\bestmeanstd}[2]{\cellcolor{bestcell}\textbf{#1} $\pm$ \scriptsize{\textbf{#2}}}
\newcommand{\bestnum}[1]{\cellcolor{bestcell}\textbf{#1}}

\usepackage{wrapfig}
\usepackage{url}

\newcommand{\meanstd}[2]{#1 $\pm$ \scriptsize{#2}}

\usepackage[pagebackref,breaklinks,colorlinks,allcolors=blue]{hyperref}

\title{HERO: A Heterogeneity-Aware Benchmark Library for Federated Continual Learning}

\author{%
{\small Thinh T. H. Nguyen\thanks{Co-first Authors.}, Le-Tuan Nguyen$^*$, Minh-Duong Nguyen, Nhi Trinh,}\\
{\small \textbf{Anh Tran Nam Nguyet, Dung D. Le$^\dagger$, Kok-Seng Wong}\thanks{Co-corresponding Authors: \url{dung.ld@vinuni.edu.vn} and \url{wong.ks@vinuni.edu.vn}.}}\\
{\normalfont\small VinUniversity, Hanoi, Vietnam}\\
{\normalfont\small \texttt{\{thinh.nth,tuan.nl,duong.nm2,23nhi.ttt,23anh.tnn2,dung.ld,wong.ks\}@vinuni.edu.vn}}
}

\begin{document}

\maketitle

\begin{abstract}
Federated continual learning (FCL) evaluates how distributed clients learn from changing data streams while retaining previously learned knowledge. Existing evaluations are difficult to compare because they often change datasets, task splits, client data splits, task orders, backbones, memory assumptions, and reporting rules simultaneously. We introduce \textbf{HERO}, a heterogeneity-aware benchmark library for FCL. HERO builds benchmark streams by separating three choices that are often coupled, namely the task split, the client data split, and the client task sequence. In HERO-Core, the main comparable benchmark, $\alpha$ controls client data skew and $\rho$ controls task-order mismatch. We evaluate representative FCL methods on CIFAR-100 and TinyImageNet using final average accuracy, average forgetting, and bottom-10\% client accuracy. We also include a graph-based Domain-IL portability case study on OGB-MolPCBA, where scaffold-domain granularity changes the input distribution while the prediction task remains fixed. Our results show that method behavior changes across easy and heterogeneous settings, that average accuracy can hide weak bottom-client performance, that task-order mismatch favors different strategies from synchronized evaluation, and that the same HERO interface can expose domain-shift difficulty beyond image-based FCIL. HERO releases benchmark streams, configurations, method implementations, and reporting scripts to support reproducible and setting-aware FCL evaluation.
\end{abstract}

\newcommand{\md}[1]{{\color{orange}{[#1]}}}
\newcommand{\err}[1]{{\color{red}{#1}}}

\section{Motivation and Benchmark Gap}
\label{sec:motivation}

Federated learning (FL) allows many clients to train a shared model without centralizing raw data, but standard FL evaluation usually treats the learning problem as fixed over time~\citep{mcmahan2017communication,kairouz2021advances}. This assumption is often too narrow in practice. Clients may collect data continuously, new classes may appear, and the model may need to adapt to new distributions without losing earlier knowledge. Federated continual learning (FCL) studies this setting by combining federated optimization with continual learning (CL)~\citep{yoon2021federated,ma2022continual,le2021federated}. This combination makes evaluation difficult. On the FL side, methods must handle non-IID client data, client drift, and partial participation~\citep{zhao2018federated,hsu2019measuring,li2020fedprox,karimireddy2020scaffold,wang2020tackling}. On the CL side, methods must handle non-stationary streams, growing task sequences, and forgetting~\citep{kirkpatrick2017overcoming,li2017learning,lopez2017gradient,rebuffi2017icarl,de2021continual,van2022three}. A common FCL setting is federated class-incremental learning (FCIL), where clients and the server learn an expanding label space without task identity at test time~\citep{dong2022federated,babakniya2023data,zhang2023target}.

Recent FCL methods use replay, generative replay, distillation, task tracing, prompt tuning, personalization, class balancing, gradient correction, and resource-aware training~\citep{babakniya2023data,zhang2023target,tran2024text,li2024towards,wang2024traceable,dong2023no,qi2025class,Li2025ResourceConstrainedFC}. These techniques address important problems, but their reported results are still hard to compare. Different papers often change the dataset, task split, client data split, task sequence, backbone, memory budget, and reporting rule at the same time. As a result, a method may look stronger because it is more robust, but it may also look stronger because it is evaluated under an easier stream or a different set of assumptions. This creates a benchmark gap. FCL needs a reusable library that can construct comparable streams, expose the main sources of heterogeneity, document method assumptions, and report more than one average score. Without such a library, a result table may hide whether a method is robust to client data skew, robust to task-sequence mismatch, robust to domain shift, or only strong under one favorable protocol.

Thefore, we introduce \textbf{HERO}, a benchmark library for FCL evaluation. HERO focuses on making evaluation more controlled and easier to interpret. The main design is to separate three choices that are often coupled in prior work. The first choice is the \emph{task split}, which decides which classes or domains form each task. The second choice is the \emph{client data split}, which decides what data each client receives for each task. The third choice is the \emph{client task sequence}, which decides when each client encounters each task. By separating these choices, HERO can study client data skew and task-sequence mismatch as different sources of difficulty.

HERO-Core is the main comparable benchmark in this paper. It focuses on image-based FCIL with CIFAR-100 and TinyImageNet, where representative methods are evaluated under shared stream construction and controlled heterogeneity settings. The core results show that method behavior changes as client streams move from easy synchronized settings to more heterogeneous settings. In particular, methods that are strong under synchronized evaluation are not always the most stable when client data skew and task-sequence mismatch increase, and average performance can hide weak behavior on the lowest-performing clients. To show that the library interface is not limited to image-based FCIL, we also include a portability case study on OGB-MolPCBA~\citep{hu2020open}. This graph-based Domain-IL study uses scaffold-grouped molecular domains and shows that finer domain granularity makes FCIL evaluation more difficult. To summarize, this paper makes four contributions:
\begin{itemize}[leftmargin=*]
\item We introduce \textbf{HERO}, a heterogeneity-aware benchmark library for FCL that unifies stream construction, method execution, result reporting, and released benchmark files.

\item To make heterogeneous FCL evaluation easier to interpret, we provide a stream generator that separates task split, client data split, and client task sequence, so that different sources of difficulty can be controlled and inspected separately.

\item To support fair and reusable comparison, we define HERO-Core, a dense image-based FCIL benchmark on CIFAR-100 and TinyImageNet with shared streams, controlled heterogeneity settings, and a representative method pool.

\item We evaluate HERO on CIFAR-100, TinyImageNet, and an OGB-MolPCBA portability case study, showing that method behavior changes across heterogeneity settings, bottom-client robustness can differ from average performance, and the same interface can extend to graph-based Domain-IL without mixing incompatible settings into one leaderboard.
\end{itemize}

\section{Related Work}
\label{sec:related}

\paragraph{Federated learning benchmarks and libraries.}
Several benchmarks and libraries have shaped FL evaluation, including LEAF for federated datasets and metrics, FedScale for scalable system evaluation, Flower for flexible FL experimentation, and FLamby for realistic cross-silo healthcare FL~\citep{caldas2018leaf,lai2022fedscale,beutel2020flower,ogier2022flamby}. These resources mainly target static FL tasks or system-level evaluation. HERO complements them by focusing on FCL, where the benchmark must construct task streams, measure forgetting, and handle client-specific task sequences. For portability beyond vision, HERO also uses OGB-MolPCBA as a graph-based Domain-IL case study, because scaffold-grouped molecular domains naturally create distribution shifts and average precision is the standard metric for its imbalanced multi-label prediction task~\citep{hu2020open}.

\paragraph{Continual and federated continual learning.}
CL studies how models learn from changing data while retaining previous knowledge, with common strategies including regularization, distillation, replay, and gradient constraints~\citep{kirkpatrick2017overcoming,li2017learning,lopez2017gradient,chaudhry2018efficient,buzzega2020dark}. Class-IL is especially difficult because the label space grows and task identity is unavailable at inference~\citep{rebuffi2017icarl,hou2019learning,wu2019large,van2022three}. FCL combines this non-stationary learning problem with decentralized optimization, and prior work has studied inter-client transfer, distillation, asynchronous streams, heterogeneous tasks, online streams, and resource constraints~\citep{yoon2021federated,ma2022continual,le2021federated,shenaj2023afcl,luopan2023fedknow,wuerkaixi2024accurate,Li2025ResourceConstrainedFC,serra2025federated}. HERO follows this line, but focuses on making the evaluation protocol controlled and reusable.

\paragraph{Federated class-incremental learning methods.}
FCIL has become a central FCL setting. Existing methods study class-aware learning, data-free or generative replay, exemplar-free distillation, text-enhanced replay, efficient replay, traceable learning, realistic FCIL, class-balanced replay, prompting, LoRA-based adaptation, gradient correction, and personalization~\citep{dong2022federated,babakniya2023data,qi2023better,nguyen2024overcoming,zhang2023target,tran2024text,li2024towards,li2025re,wang2024traceable,dong2023no,qi2025class,guo2024pilora,piao2024federated,salami2025closed,zhang2025pfedmxf,zhang2025fedagc}. This active method landscape makes unified evaluation important, because conclusions can change with the stream construction, heterogeneity setting, and reported metric. A fuller discussion of related work and method categories is provided in Appendix~\ref{app:full_related}.

\section{Benchmark Design and Stream Construction}
\label{sec:design_construction}

\textbf{HERO} is a benchmark library for FCL, not a single fixed leaderboard. The library has three tiers. \textbf{HERO-Core} is the main comparable benchmark used in this paper, and it focuses on image-based FCIL. \textbf{HERO-Portability} tests whether the same interface extends to other settings, such as domain-incremental learning (Domain-IL), text streams, and graph streams. \textbf{HERO-Extended} stores additional dataset constructors, scenario builders, and method implementations for future use. This tiered design lets the library grow without mixing incompatible datasets, losses, backbones, and metrics into one score. The full dataset registry is provided in Appendix~\ref{app:dataset_registry}.

\subsection{Generated benchmark stream}

HERO first generates a benchmark stream, then reuses the same stream for every method. In a class-incremental learning (Class-IL) setting, the global label space is split into $T$ disjoint class sets,
\begin{equation}
\mathcal{Y}
=
\bigcup_{t=1}^{T} \mathcal{C}_t,
\qquad
\mathcal{C}_i \cap \mathcal{C}_j = \emptyset
\quad \text{for } i \neq j .
\label{eq:task_partition}
\end{equation}
Each task contains the samples whose labels belong to its class set. A generated HERO stream is written as
\begin{equation}
\mathcal{B}(\xi,\alpha,\rho)
=
\left\{
\mathcal{C}_{1:T},
\mathcal{D}^{\mathrm{tr}}_{k,t},
\mathcal{D}^{\mathrm{te}}_{k,t},
\pi_{1:K}
\right\}.
\label{eq:benchmark_stream}
\end{equation}
Here, $\xi$ is the benchmark seed, $\mathcal{C}_{1:T}$ is the task split, $\mathcal{D}^{\mathrm{tr}}_{k,t}$ and $\mathcal{D}^{\mathrm{te}}_{k,t}$ are the train and test shards of client $k$ for task $t$, and $\pi_k$ is the task sequence of client $k$. The same seed also fixes the client sampling schedule and evaluation protocol. These released files make the stream reproducible, while the compact notation keeps the main construction easy to read.

In Domain-IL, the same interface is used, but tasks correspond to domains or distribution groups instead of new class sets. The output space stays fixed, while the input distribution changes across tasks. Full construction details are provided in Appendix~\ref{app:generator_details}.

\subsection{Three construction steps}

HERO separates stream construction into three steps. First, it creates the task split. Second, it creates the client data split. Third, it creates the client task sequence. This separation is important because client data skew and task-order mismatch should be controlled independently, rather than changed together by one opaque benchmark protocol.
The client data split is controlled by $\alpha$. For each task $t$ and each class $c \in \mathcal{C}_t$, HERO samples client proportions as
\begin{equation}
\mathbf{p}_{t,c}
\sim
\mathrm{Dir}_K(\alpha \mathbf{1}).
\label{eq:dirichlet_split}
\end{equation}
Smaller $\alpha$ creates stronger client data skew, while larger $\alpha$ creates more balanced client data. The split is applied inside each task, so every client eventually receives data from every task in HERO-Core, although the amount and class proportions may differ.
The client task sequence is controlled by $\rho$. HERO starts from the shared order
\begin{equation}
\pi^\star = (1,2,\ldots,T).
\label{eq:reference_order}
\end{equation}
When $\rho=0$, all clients follow this order. Larger $\rho$ values make client task sequences more different. Changing $\rho$ only reorders already generated client-task shards, so it does not change the task split or the client data split.

\subsection{Construction checks}

Nominal settings may not fully describe the generated stream because finite data, integer rounding, and permutation sampling can change the realized difficulty. HERO therefore checks each stream after generation. The checks report the realized client data skew, the pairwise mismatch among client task sequences, and the mismatch from the shared reference order. In the main experiments, these checks verify that smaller $\alpha$ creates stronger data skew and larger $\rho$ creates stronger task-order mismatch. Full definitions are provided in Appendix~\ref{app:generator_details}.

\subsection{Metrics and released files}

HERO-Core reports three metrics. Final average accuracy (AFA) measures final performance across clients and tasks. Average forgetting (AF) measures how much old-task performance drops by the end of training. Bottom-10\% client accuracy (B10) measures final performance on the lowest-performing clients. Higher AFA and B10 are better, while lower AF is better. Portability studies may use task-specific metrics when accuracy is not appropriate. For OGB-MolPCBA, HERO reports Average Precision (AP), because the dataset is a highly imbalanced multi-label molecular property prediction benchmark. Its forgetting score is measured in AP points. Full metric definitions are provided in Appendix~\ref{app:metrics}. HERO releases each stream as reusable files, including the configuration, task split, client data split, client task sequence, client sampling schedule, test split, raw results, and report scripts. This separates stream generation from method execution. A new method can be evaluated by loading the same stream, and a new dataset can be added by producing the same file format. The full release format is provided in Appendix~\ref{app:artifact_schema}.

\section{Baseline Experiments and Benchmark Utility}
\label{sec:experiments}

This section uses HERO-Core to test whether the benchmark provides useful measurement signals. We first evaluate image-based FCIL on CIFAR-100 and TinyImageNet, where all methods share the same stream construction, training budget, and metric set. We then add one portability case study on OGB-MolPCBA to test whether the same stream interface can support graph-based Domain-IL. The goal is therefore not only to rank methods, but also to check whether HERO separates method behavior, exposes bottom-client failures, and extends beyond the core image setting without mixing incompatible results into one leaderboard.

\subsection{Anchor-Setting Results}

We compare methods in three anchor settings. The easy setting is $(\alpha=100.00,\rho=0.00)$, where client data skew is weak and all clients follow the same task sequence. The mild setting is $(\alpha=10.00,\rho=0.50)$, where both client data skew and task-order mismatch are moderate. The joint-hard setting is $(\alpha=0.10,\rho=1.00)$, where both sources of heterogeneity are strong. These settings form an interpretable path from synchronized evaluation to strongly heterogeneous evaluation. The two single-axis settings, alpha-hard and rho-hard, are reported in Appendix~\ref{app:anchor_results}.

Tables~\ref{tab:anchor_cifar_main} and~\ref{tab:anchor_tiny_main} show that HERO-Core is discriminative. Specialized FCIL methods consistently outperform Local-Only and FedAvg, which confirms that the benchmark is not only measuring standard federated optimization. More importantly, the mild setting already changes the relative behavior of methods. FedCBDR is strongest in the easy setting, but TagFed and LGA become more competitive as task-order mismatch and client data skew increase, especially on B10. This supports the main motivation of HERO. A synchronized leaderboard is useful, but it is not sufficient to describe method behavior under heterogeneous client streams.

\begin{table}[t]
\centering
\caption{Anchor-setting results on CIFAR-100, reported as mean $\pm$ standard deviation over five seeds. Easy is $(\alpha=100.00,\rho=0.00)$, mild is $(\alpha=10.00,\rho=0.50)$, and joint-hard is $(\alpha=0.10,\rho=1.00)$. Highlighted cells indicate the best value in each column. Higher is better for AFA and B10. Lower is better for AF.}
\label{tab:anchor_cifar_main}
\renewcommand\arraystretch{1.12}
\footnotesize
\setlength{\tabcolsep}{3pt}
\resizebox{\textwidth}{!}{
\begin{tabular}{lccc ccc ccc}
\toprule
\multirow{2}{*}{Method}
& \multicolumn{3}{c}{Easy}
& \multicolumn{3}{c}{Mild}
& \multicolumn{3}{c}{Joint-hard} \\
\cmidrule(lr){2-4}\cmidrule(lr){5-7}\cmidrule(lr){8-10}
& AFA $\uparrow$ & AF $\downarrow$ & B10 $\uparrow$
& AFA $\uparrow$ & AF $\downarrow$ & B10 $\uparrow$
& AFA $\uparrow$ & AF $\downarrow$ & B10 $\uparrow$ \\
\midrule
Local-Only & \meanstd{34.21}{2.13} & \meanstd{55.37}{2.23} & \meanstd{31.83}{2.31} & \meanstd{29.14}{2.31} & \meanstd{62.43}{2.37} & \meanstd{22.76}{2.13} & \meanstd{20.73}{2.63} & \meanstd{70.42}{2.53} & \meanstd{12.41}{1.83} \\
FedAvg     & \meanstd{37.43}{2.43} & \meanstd{57.19}{2.31} & \meanstd{34.52}{2.33} & \meanstd{31.82}{2.51} & \meanstd{64.91}{2.43} & \meanstd{24.51}{2.23} & \meanstd{22.69}{2.93} & \meanstd{72.83}{2.63} & \meanstd{13.16}{2.03} \\
GLFC       & \meanstd{50.11}{1.83} & \meanstd{34.27}{1.73} & \meanstd{47.62}{1.93} & \meanstd{43.62}{2.03} & \meanstd{40.58}{1.83} & \meanstd{35.74}{2.13} & \meanstd{32.57}{2.53} & \meanstd{50.61}{2.13} & \meanstd{22.47}{2.13} \\
MFCL       & \meanstd{44.83}{1.93} & \meanstd{32.61}{1.63} & \meanstd{41.39}{2.03} & \meanstd{39.27}{2.03} & \meanstd{37.92}{1.73} & \meanstd{32.63}{2.23} & \meanstd{31.83}{2.43} & \meanstd{47.92}{2.03} & \meanstd{22.83}{2.03} \\
TARGET     & \meanstd{32.79}{1.83} & \meanstd{36.43}{1.73} & \meanstd{30.12}{1.93} & \meanstd{28.61}{1.93} & \meanstd{42.53}{1.83} & \meanstd{23.84}{2.13} & \meanstd{25.42}{2.33} & \meanstd{51.33}{2.13} & \meanstd{18.92}{2.03} \\
LANDER     & \meanstd{43.21}{1.73} & \meanstd{25.63}{1.43} & \meanstd{40.77}{1.83} & \meanstd{37.82}{1.91} & \meanstd{31.97}{1.63} & \meanstd{31.92}{2.11} & \meanstd{33.76}{2.23} & \meanstd{38.71}{1.83} & \meanstd{27.63}{1.93} \\
Re-Fed+    & \meanstd{47.89}{1.83} & \meanstd{28.17}{1.53} & \meanstd{44.36}{1.93} & \meanstd{41.73}{1.93} & \meanstd{34.81}{1.73} & \meanstd{34.73}{2.13} & \meanstd{36.21}{2.23} & \meanstd{40.52}{1.93} & \meanstd{28.41}{1.93} \\
TagFed     & \meanstd{50.73}{1.63} & \meanstd{24.61}{1.33} & \meanstd{47.94}{1.83} & \meanstd{47.92}{1.83} & \bestmeanstd{29.63}{1.53} & \bestmeanstd{42.71}{1.83} & \bestmeanstd{42.83}{2.13} & \bestmeanstd{34.27}{1.83} & \bestmeanstd{35.92}{1.73} \\
LGA        & \meanstd{49.67}{1.73} & \meanstd{25.41}{1.33} & \meanstd{46.82}{1.83} & \meanstd{46.87}{1.83} & \meanstd{30.27}{1.53} & \meanstd{41.13}{1.93} & \meanstd{41.96}{2.23} & \meanstd{35.43}{1.83} & \meanstd{34.71}{1.83} \\
FedCBDR    & \bestmeanstd{53.91}{1.53} & \bestmeanstd{22.73}{1.23} & \bestmeanstd{50.86}{1.73} & \bestmeanstd{48.16}{1.73} & \meanstd{31.72}{1.63} & \meanstd{40.93}{1.83} & \meanstd{39.74}{2.03} & \meanstd{39.93}{1.93} & \meanstd{31.87}{1.73} \\
\bottomrule
\end{tabular}}
\end{table}

\begin{table}[t]
\centering
\caption{Anchor-setting results on TinyImageNet, reported as mean $\pm$ standard deviation over five seeds. Easy is $(\alpha=100.00,\rho=0.00)$, mild is $(\alpha=10.00,\rho=0.50)$, and joint-hard is $(\alpha=0.10,\rho=1.00)$. Highlighted cells indicate the best value in each column. Higher is better for AFA and B10. Lower is better for AF.}
\label{tab:anchor_tiny_main}
\renewcommand\arraystretch{1.12}
\footnotesize
\setlength{\tabcolsep}{3pt}
\resizebox{\textwidth}{!}{
\begin{tabular}{lccc ccc ccc}
\toprule
\multirow{2}{*}{Method}
& \multicolumn{3}{c}{Easy}
& \multicolumn{3}{c}{Mild}
& \multicolumn{3}{c}{Joint-hard} \\
\cmidrule(lr){2-4}\cmidrule(lr){5-7}\cmidrule(lr){8-10}
& AFA $\uparrow$ & AF $\downarrow$ & B10 $\uparrow$
& AFA $\uparrow$ & AF $\downarrow$ & B10 $\uparrow$
& AFA $\uparrow$ & AF $\downarrow$ & B10 $\uparrow$ \\
\midrule
Local-Only & \meanstd{21.46}{1.83} & \meanstd{58.43}{2.13} & \meanstd{18.72}{1.63} & \meanstd{17.63}{1.93} & \meanstd{65.28}{2.23} & \meanstd{11.94}{1.33} & \meanstd{12.84}{2.23} & \meanstd{73.26}{2.53} & \meanstd{5.13}{1.13} \\
FedAvg     & \meanstd{22.83}{1.93} & \meanstd{60.17}{2.23} & \meanstd{19.31}{1.63} & \meanstd{18.72}{2.03} & \meanstd{67.53}{2.33} & \meanstd{12.36}{1.43} & \meanstd{13.31}{2.33} & \meanstd{75.84}{2.63} & \meanstd{5.62}{1.23} \\
GLFC       & \meanstd{35.47}{1.63} & \meanstd{38.62}{1.73} & \meanstd{32.76}{1.53} & \meanstd{29.14}{1.83} & \meanstd{45.92}{1.93} & \meanstd{20.83}{1.43} & \meanstd{22.17}{2.13} & \meanstd{55.92}{2.23} & \meanstd{11.82}{1.33} \\
MFCL       & \meanstd{30.72}{1.63} & \meanstd{36.91}{1.63} & \meanstd{27.83}{1.53} & \meanstd{25.91}{1.83} & \meanstd{43.67}{1.83} & \meanstd{18.92}{1.43} & \meanstd{20.73}{2.13} & \meanstd{52.71}{2.13} & \meanstd{11.16}{1.33} \\
TARGET     & \meanstd{27.31}{1.53} & \meanstd{31.72}{1.53} & \meanstd{25.54}{1.43} & \meanstd{23.64}{1.73} & \meanstd{38.27}{1.73} & \meanstd{17.83}{1.43} & \meanstd{19.82}{2.03} & \meanstd{48.63}{2.03} & \meanstd{10.94}{1.33} \\
LANDER     & \meanstd{31.29}{1.53} & \bestmeanstd{23.63}{1.33} & \meanstd{29.52}{1.43} & \meanstd{27.16}{1.73} & \bestmeanstd{30.83}{1.63} & \meanstd{22.31}{1.43} & \meanstd{23.91}{1.93} & \meanstd{39.84}{1.93} & \meanstd{14.37}{1.33} \\
Re-Fed+    & \meanstd{32.71}{1.53} & \meanstd{28.53}{1.43} & \meanstd{30.14}{1.43} & \meanstd{28.42}{1.73} & \meanstd{34.19}{1.63} & \meanstd{22.72}{1.43} & \meanstd{24.52}{1.93} & \meanstd{43.17}{1.93} & \meanstd{14.26}{1.33} \\
TagFed     & \meanstd{36.83}{1.43} & \meanstd{27.31}{1.33} & \meanstd{34.91}{1.33} & \bestmeanstd{33.62}{1.63} & \meanstd{32.87}{1.53} & \bestmeanstd{28.96}{1.33} & \bestmeanstd{29.41}{1.83} & \bestmeanstd{38.92}{1.83} & \bestmeanstd{21.83}{1.23} \\
LGA        & \meanstd{35.72}{1.43} & \meanstd{28.72}{1.33} & \meanstd{33.61}{1.33} & \meanstd{32.81}{1.63} & \meanstd{33.94}{1.53} & \meanstd{27.91}{1.33} & \meanstd{28.76}{1.83} & \meanstd{40.14}{1.83} & \meanstd{20.96}{1.23} \\
FedCBDR    & \bestmeanstd{38.24}{1.33} & \meanstd{25.92}{1.23} & \bestmeanstd{35.83}{1.23} & \meanstd{32.67}{1.53} & \meanstd{35.61}{1.43} & \meanstd{26.82}{1.23} & \meanstd{27.42}{1.73} & \meanstd{43.71}{1.83} & \meanstd{18.52}{1.23} \\
\bottomrule
\end{tabular}}
\end{table}

\begin{figure*}[t]
\centering
\includegraphics[width=\textwidth]{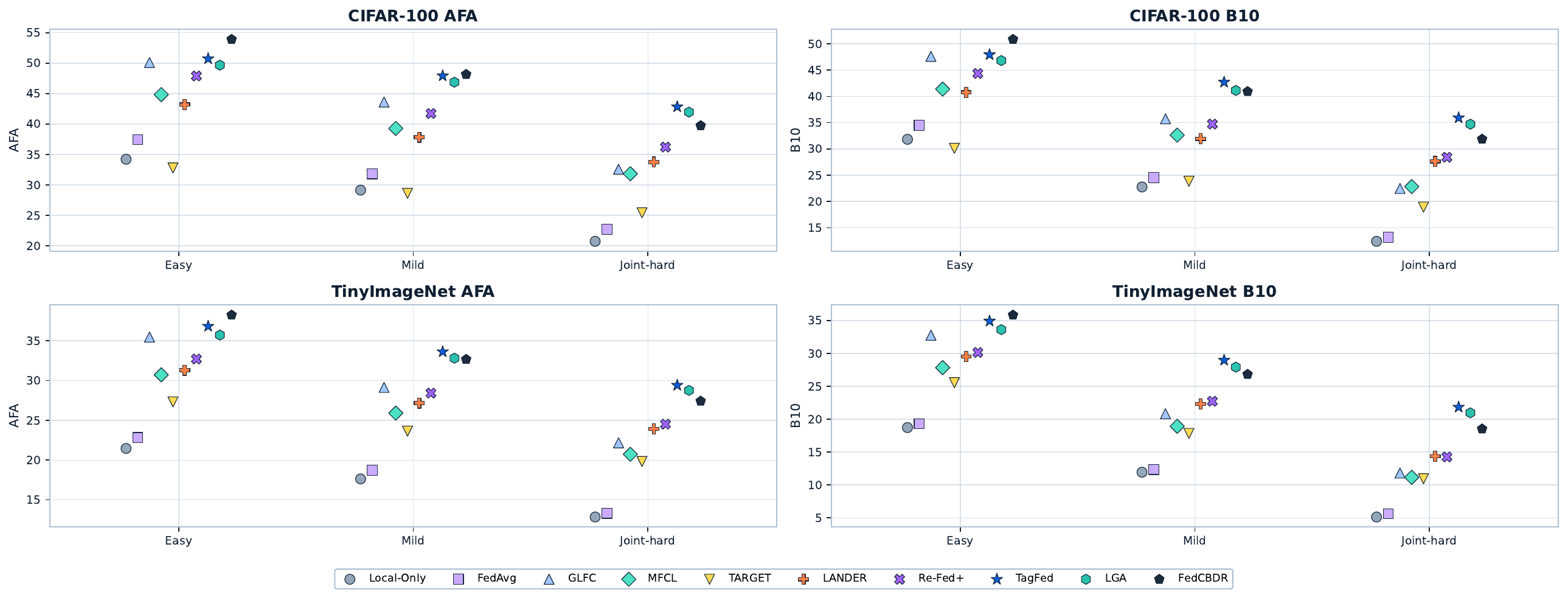}
\caption{
Anchor-setting profiles from easy to mild to joint-hard settings. Each point represents one method, and the panels show AFA and B10 on CIFAR-100 and TinyImageNet. The figure shows that degradation is not uniform across methods, and that the strongest method under synchronized evaluation is not always the most stable method under stronger heterogeneity.
}
\label{fig:regime_degradation_profiles}
\end{figure*}

\subsection{Full-Grid Robustness Summary}

\begin{figure*}[t]
\centering
\includegraphics[width=\textwidth]{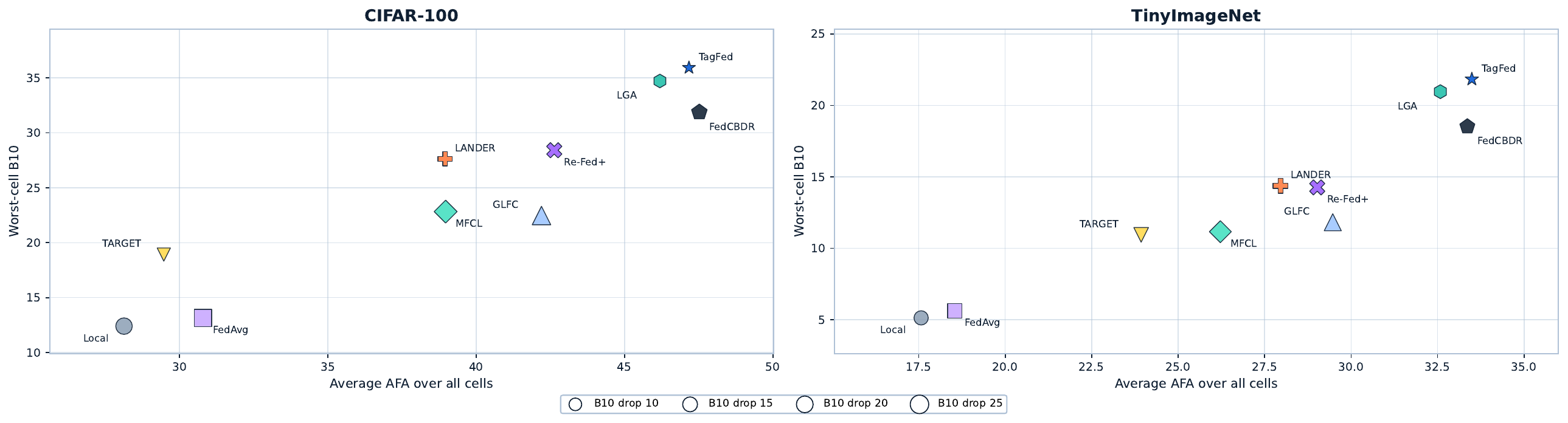}
\caption{
Full-grid robustness summary. Each point represents one method. The horizontal axis measures average AFA across the full $\alpha$-$\rho$ grid, while the vertical axis measures worst-cell B10. Bubble size indicates the B10 drop from easy to joint-hard.
}
\label{fig:full_grid_robustness}
\end{figure*}

Anchor settings give interpretable snapshots, while the full $\alpha$-$\rho$ grid tests whether the same pattern holds across all benchmark cells. We therefore aggregate each method over the full $3\times3$ grid. Table~\ref{tab:full_grid_robustness} reports average AFA, worst-cell AFA, AFA drop, average B10, worst-cell B10, and B10 drop. The average score measures overall performance, the worst-cell score measures robustness under the hardest generated cell, and the drop measures degradation from easy to joint-hard. Figure~\ref{fig:full_grid_robustness} gives the same information visually by placing average AFA and worst-cell B10 in one plot.

The full-grid summary confirms the anchor-setting story. FedCBDR remains strong in average performance, but its easy-to-joint drop is larger than that of TagFed and LGA. In contrast, TagFed and LGA have stronger worst-cell B10, which means they better protect difficult clients under severe heterogeneity. Thus, average performance and bottom-client robustness are related but not identical. This is why HERO reports AFA, AF, and B10 together instead of relying on one global score.

\begin{table}[t]
\centering
\caption{Full-grid method robustness summary. Grid mean is averaged over all $3\times3$ $\alpha$-$\rho$ cells. Worst-cell is the lowest score over the grid. Drop is the decrease from the easy setting to the joint-hard setting. Highlighted cells indicate the best value in each column according to the column direction. Higher is better for grid mean and worst-cell scores. Lower is better for drops.}
\label{tab:full_grid_robustness}
\renewcommand\arraystretch{1.10}
\footnotesize
\setlength{\tabcolsep}{3pt}
\begin{tabular}{llcccccc}
\toprule
Dataset & Method 
& Avg. AFA $\uparrow$ 
& Worst AFA $\uparrow$ 
& AFA Drop $\downarrow$
& Avg. B10 $\uparrow$
& Worst B10 $\uparrow$
& B10 Drop $\downarrow$ \\
\midrule
\multirow{10}{*}{CIFAR-100}
& Local-Only & 28.13 & 20.73 & 13.48 & 23.08 & 12.41 & 19.42 \\
& FedAvg     & 30.79 & 22.69 & 14.74 & 24.89 & 13.16 & 21.36 \\
& GLFC       & 42.20 & 32.57 & 17.54 & 36.28 & 22.47 & 25.15 \\
& MFCL       & 38.97 & 31.83 & 13.00 & 33.02 & 22.83 & 18.56 \\
& TARGET     & 29.47 & 25.42 & \bestnum{7.37} & 25.07 & 18.92 & \bestnum{11.20} \\
& LANDER     & 38.95 & 33.76 & 9.45  & 34.85 & 27.63 & 13.14 \\
& Re-Fed+    & 42.63 & 36.21 & 11.68 & 37.17 & 28.41 & 15.95 \\
& TagFed     & 47.17 & \bestnum{42.83} & 7.90 & \bestnum{42.52} & \bestnum{35.92} & 12.02 \\
& LGA        & 46.19 & 41.96 & 7.71  & 41.36 & 34.71 & 12.11 \\
& FedCBDR    & \bestnum{47.52} & 39.74 & 14.17 & 42.31 & 31.87 & 18.99 \\
\midrule
\multirow{10}{*}{TinyImageNet}
& Local-Only & 17.57 & 12.84 & 8.62  & 12.59 & 5.13  & 13.59 \\
& FedAvg     & 18.54 & 13.31 & 9.52  & 13.14 & 5.62  & 13.69 \\
& GLFC       & 29.47 & 22.17 & 13.30 & 23.32 & 11.82 & 20.94 \\
& MFCL       & 26.22 & 20.73 & 9.99  & 20.32 & 11.16 & 16.67 \\
& TARGET     & 23.93 & 19.82 & 7.49  & 18.96 & 10.94 & 14.60 \\
& LANDER     & 27.96 & 23.91 & 7.38  & 22.69 & 14.37 & 15.15 \\
& Re-Fed+    & 29.02 & 24.52 & 8.19  & 22.98 & 14.26 & 15.88 \\
& TagFed     & \bestnum{33.49} & \bestnum{29.41} & 7.42 & \bestnum{29.01} & \bestnum{21.83} & 13.08 \\
& LGA        & 32.58 & 28.76 & \bestnum{6.96} & 27.91 & 20.96 & \bestnum{12.65} \\
& FedCBDR    & 33.36 & 27.42 & 10.82 & 28.03 & 18.52 & 17.31 \\
\bottomrule
\end{tabular}
\end{table}

\subsection{Portability Case Study on OGB-MolPCBA}

The previous results evaluate HERO-Core, where all methods are compared under the same image-based FCIL protocol. To test whether the same stream interface can support a different setting, we add a graph-based Domain-IL case study on OGB-MolPCBA~\citep{hu2020open}. This study is not merged into the HERO-Core leaderboard because it uses graph inputs, fixed multi-label prediction, and AP-based evaluation. Instead, it tests extensibility under a modality and metric that differ from image-based FCIL.

\begin{figure}[t]
\centering
\includegraphics[width=0.92\linewidth]{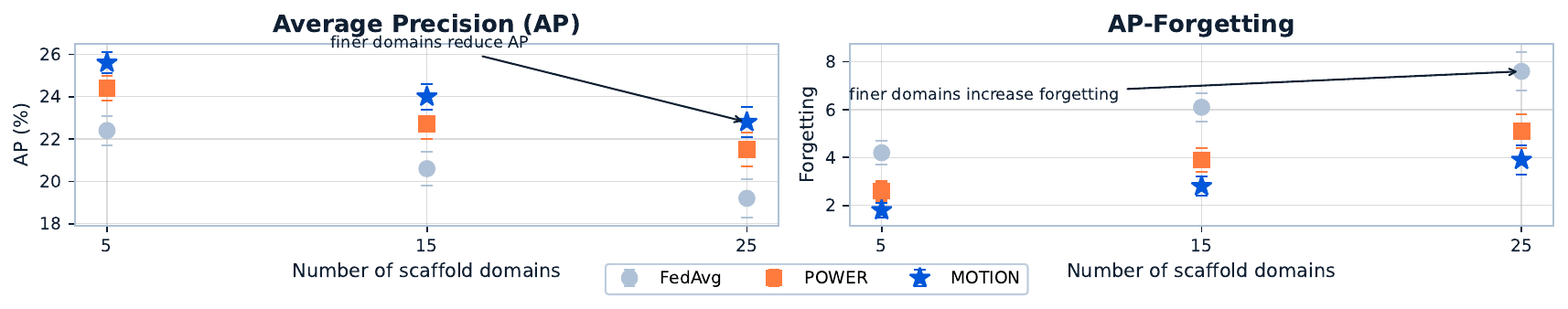}
\caption{
Graph-based Domain-IL portability on OGB-MolPCBA. Increasing the number of scaffold domains reduces AP and increases AP-forgetting, while POWER and MOTION remain more stable than FedAvg. This case study is reported separately from HERO-Core because it uses graph inputs, fixed multi-label prediction, and AP-based evaluation.
}
\label{fig:ogb_domain_granularity}
\end{figure}

We construct Domain-IL streams by grouping molecular scaffolds into $G$ ordered domains, with $G\in\{5,15,25\}$. The prediction space stays fixed with $128$ binary tasks, but the input distribution changes as the model moves across scaffold groups. This makes OGB-MolPCBA a useful portability test. It checks whether HERO can represent distribution-shift streams where the challenge is not new classes, but changing molecular domains. Since most image-based FCIL methods are not designed for graph topology or graph-level multi-label prediction, we compare FedAvg with two graph-based FCL methods, POWER and MOTION~\citep{zhu2025federated,wan2025motion}. Table~\ref{tab:ogb_portability_main} and Figure~\ref{fig:ogb_domain_granularity} show that AP decreases as the number of scaffold domains increases, while AP-forgetting increases. This trend means that finer scaffold grouping creates sharper domain transitions and stronger interference across domains. The result supports the library claim of HERO. The same stream and reporting interface can describe both image-based FCIL and graph-based Domain-IL, while the tiered organization prevents these incompatible settings from being collapsed into one leaderboard.

\begin{table}[t]
\centering
\caption{
OGB-MolPCBA Domain-IL portability case study. We report mean $\pm$ standard deviation over five seeds. 
$G$ denotes the number of scaffold-grouped domains. AP is reported in percent, and forgetting is measured in AP points. 
Highlighted cells indicate the best value for each scaffold-domain setting. Higher AP is better, while lower forgetting is better.
}
\label{tab:ogb_portability_main}
\renewcommand\arraystretch{1.10}
\footnotesize
\setlength{\tabcolsep}{3pt}
\begin{tabular}{lcc cc cc}
\toprule
\multirow{2}{*}{Method}
& \multicolumn{2}{c}{$G=5$}
& \multicolumn{2}{c}{$G=15$}
& \multicolumn{2}{c}{$G=25$} \\
\cmidrule(lr){2-3}\cmidrule(lr){4-5}\cmidrule(lr){6-7}
& AP $\uparrow$ & Forgetting $\downarrow$
& AP $\uparrow$ & Forgetting $\downarrow$
& AP $\uparrow$ & Forgetting $\downarrow$ \\
\midrule
FedAvg
& \meanstd{22.40}{0.70} & \meanstd{4.20}{0.50}
& \meanstd{20.60}{0.80} & \meanstd{6.10}{0.60}
& \meanstd{19.20}{0.90} & \meanstd{7.60}{0.80} \\

POWER~\citep{zhu2025federated}
& \meanstd{24.40}{0.60} & \meanstd{2.60}{0.40}
& \meanstd{22.70}{0.70} & \meanstd{3.90}{0.50}
& \meanstd{21.50}{0.80} & \meanstd{5.10}{0.70} \\

MOTION~\citep{wan2025motion}
& \bestmeanstd{25.60}{0.50} & \bestmeanstd{1.80}{0.30}
& \bestmeanstd{24.00}{0.60} & \bestmeanstd{2.80}{0.40}
& \bestmeanstd{22.80}{0.70} & \bestmeanstd{3.90}{0.60} \\
\bottomrule
\end{tabular}
\end{table}

Sanity checks for method separation, seed stability, and AFA-B10 gap are reported in Appendix~\ref{app:sanity_checks}. We move them to the appendix because they support the reliability of the benchmark, while the main text focuses on the core empirical story and the portability case study.

\section{What HERO Reveals About Heterogeneity}
\label{sec:findings}

This section summarizes what HERO reveals about heterogeneity after the benchmark utility results. We keep the analysis tied to the same metrics, AFA, AF, and B10, instead of adding extra ranking metrics. The goal is to understand whether the generated streams match the intended difficulty, whether client data skew and task-order mismatch cause different failures, and whether method choice changes across settings.

\begin{table}[t]
\centering
\caption{Construction checks for HERO-Core. Exact zeros at $\rho=0.00$ are by construction. Values are averaged over core datasets and five seeds. Full definitions of these construction checks are provided in Appendix~\ref{app:generator_details}.}
\label{tab:construction_check}
\renewcommand\arraystretch{1.10}
\footnotesize
\begin{tabular}{cc|ccc}
\toprule
$\alpha$ & Client-skew check 
& $\rho$ & Pairwise order mismatch & Reference-order mismatch \\
\midrule
$100.00$ & \meanstd{0.01}{0.01} & $0.00$ & \meanstd{0.00}{0.00} & \meanstd{0.00}{0.00} \\
$10.00$  & \meanstd{0.03}{0.01} & $0.50$ & \meanstd{0.25}{0.03} & \meanstd{0.21}{0.02} \\
$0.10$   & \meanstd{0.31}{0.03} & $1.00$ & \meanstd{0.49}{0.03} & \meanstd{0.45}{0.03} \\
\bottomrule
\end{tabular}
\end{table}

\paragraph{Generated streams match the intended difficulty.}
The first check asks whether the nominal settings create the intended stream difficulty. Smaller $\alpha$ should produce stronger client data skew, while larger $\rho$ should produce stronger task-order mismatch. Table~\ref{tab:construction_check} shows that the generated streams follow this pattern. Client skew increases as $\alpha$ decreases, and both order-mismatch checks increase as $\rho$ increases. This matters because benchmark settings should not be specified only by nominal hyperparameters. Users should also see what kind of stream was actually generated.

\paragraph{Client data skew and task-order mismatch affect metrics differently.}
The second finding is that client data skew and task-order mismatch create different failure patterns. Table~\ref{tab:effect_signature} reports average changes on CIFAR-100 relative to the easy setting. Mild combines moderate client data skew and moderate task-order mismatch. Alpha-hard isolates client data skew, rho-hard isolates task-order mismatch, and joint-hard combines both.

\begin{wraptable}{r}{0.56\textwidth}
\vspace{-0.5em}
\centering
\caption{Average effect signatures on CIFAR-100 relative to the easy setting $(\alpha=100.00,\rho=0.00)$. Mild is $(\alpha=10.00,\rho=0.50)$, alpha-hard is $(\alpha=0.10,\rho=0.00)$, rho-hard is $(\alpha=100.00,\rho=1.00)$, and joint-hard is $(\alpha=0.10,\rho=1.00)$.}
\label{tab:effect_signature}
\renewcommand\arraystretch{1.08}
\footnotesize
\begin{tabular}{lccc}
\toprule
Setting shift & $\Delta$AFA & $\Delta$AF & $\Delta$B10 \\
\midrule
Mild       & \meanstd{-4.98}{1.31}  & \meanstd{+6.44}{1.27}  & \meanstd{-8.53}{2.14} \\
Alpha-hard & \meanstd{-7.41}{1.46}  & \meanstd{+8.79}{1.62}  & \meanstd{-11.39}{2.36} \\
Rho-hard   & \meanstd{-8.62}{1.91}  & \meanstd{+11.47}{1.91} & \meanstd{-8.24}{2.04} \\
Joint-hard & \meanstd{-11.70}{3.47} & \meanstd{+13.96}{2.59} & \meanstd{-16.79}{4.66} \\
\bottomrule
\end{tabular}
\vspace{-1em}
\end{wraptable}

Alpha-hard reduces B10 more strongly than rho-hard, which means that difficult clients are especially affected by client data skew. In contrast, rho-hard increases AF more strongly, which suggests that task-order mismatch creates more temporal interference. Joint-hard hurts all three metrics. These patterns justify separating client data split from client task sequence. If both factors are merged into one generic heterogeneous setting, the source of failure becomes harder to interpret. Per-method sensitivity tables are reported in Appendix~\ref{app:axis_sensitivity}.

\paragraph{Method choice changes across settings.}
The third finding is that method choice changes across settings, even when we only use AFA, AF, and B10. Table~\ref{tab:best_methods} summarizes the best method under easy, mild, and joint-hard settings. The easy setting favors FedCBDR on AFA and B10. The mild setting already shows a transition, where TagFed becomes strongest on bottom-client performance. Under joint-hard heterogeneity, TagFed is strongest on AFA and B10 on both datasets.

\begin{table}[t]
\centering
\caption{Best methods under easy, mild, and joint-hard settings according to the three main metrics. Highlighted cells mark the setting-wise best method for each metric. Higher is better for AFA and B10. Lower is better for AF.}
\label{tab:best_methods}
\renewcommand\arraystretch{1.10}
\footnotesize
\begin{tabular}{l l c c c}
\toprule
Dataset & Setting & Best AFA & Best AF & Best B10 \\
\midrule
CIFAR-100 & Easy & FedCBDR $(53.91)$ & FedCBDR $(22.73)$ & FedCBDR $(50.86)$ \\
CIFAR-100 & Mild & FedCBDR $(48.16)$ & TagFed $(29.63)$ & TagFed $(42.71)$ \\
CIFAR-100 & Joint-hard & TagFed $(42.83)$ & TagFed $(34.27)$ & TagFed $(35.92)$ \\
TinyImageNet & Easy & FedCBDR $(38.24)$ & LANDER $(23.63)$ & FedCBDR $(35.83)$ \\
TinyImageNet & Mild & TagFed $(33.62)$ & LANDER $(30.83)$ & TagFed $(28.96)$ \\
TinyImageNet & Joint-hard & TagFed $(29.41)$ & TagFed $(38.92)$ & TagFed $(21.83)$ \\
\bottomrule
\end{tabular}
\end{table}

This table shows why HERO reports multiple settings instead of one global leaderboard. A method that is strongest when client streams are synchronized may not remain strongest when clients follow different task sequences. It also shows why B10 is useful. AFA measures average final performance, but B10 reveals whether difficult clients are protected under stronger heterogeneity. Additional per-cell results are provided in Appendix~\ref{app:full_grid_results}.

The OGB-MolPCBA case study complements this finding from another angle. In graph-based Domain-IL, increasing scaffold-domain granularity decreases AP and increases forgetting, as shown in Table~\ref{tab:ogb_portability_main} and Figure~\ref{fig:ogb_domain_granularity}. Thus, heterogeneity does not only change method behavior in image-based FCIL. It also changes stream difficulty in non-vision Domain-IL, which supports HERO's tiered benchmark design.

\section{Release, Reproducibility, and Maintenance}
\label{sec:release}

\textbf{HERO} is designed as a reusable benchmark library rather than a one-time experimental protocol. We release benchmark streams, configuration files, dataset constructors, method implementations, metric scripts, and reporting scripts. Each HERO-Core stream can be regenerated from the dataset name, scenario name, seed, $\alpha$, and $\rho$. 
The release separates stream generation from method execution. Users can evaluate a new method by loading the same task split, client data split, client task sequence, client sampling schedule, and test split. This prevents hidden stream changes from being mistaken for algorithmic improvements. The complete file format is provided in Appendix~\ref{app:artifact_schema}.

\begin{table}[t]
\centering
\caption{Released files in \textbf{HERO}. The core release supports reproduction of the main results. The extended release supports future studies and portability experiments.}
\label{tab:release_files}
\renewcommand\arraystretch{1.12}
\footnotesize
\begin{tabular}{l p{0.51\linewidth} l}
\toprule
Released file group & Content & Tier \\
\midrule
Stream generator & Task split, client data split, and client task sequence generation & Core, Extended \\
Fixed streams & Saved task splits, client data splits, client task sequences & Core \\
Dataset registry & Core, portability, and extended dataset metadata & Core, Extended \\
Method collection & Reference baselines and implemented FCL methods & Core, Extended \\
Metric scripts & AFA, AF, B10, AP, and task-specific portability metrics & Core, Extended \\
Report scripts & Tables, full-grid summaries, and setting-wise reports & Core \\
Configuration files & YAML files for HERO-Core settings and method runs & Core \\
Portability configs & Configuration files for Domain-IL, text, and graph studies & Extended \\
\bottomrule
\end{tabular}
\end{table}

\paragraph{Versioning.}
We maintain separate versions for stream construction and method implementation. Changes to dataset constructors, task splits, client data splits, task-order samplers, or metrics create a new benchmark version. Changes to method implementation create a new method version. This separation allows users to know whether two results are directly comparable.

\paragraph{Responsible extension.}
New datasets and methods can be added to HERO, but they must declare their assumptions. For methods, this includes replay memory, generators, pretrained modules, personalization, and additional server information. For datasets, this includes modality, scenario, task construction, and metric.

\section{Discussion and Limitations}
\label{sec:discussion}

\paragraph{Discussion.}
\textbf{HERO} changes FCL evaluation by separating stream construction, method execution, and metric reporting. This separation makes results easier to interpret, because a score can be traced to the task split, the client data split, the client task sequence, and the metric used for reporting. The core results show that AFA, AF, and B10 are complementary. FedCBDR is strong under easier synchronized streams, while TagFed and LGA are more robust on B10 under stronger heterogeneity, which indicates that average performance can hide failures on difficult clients. The OGB-MolPCBA case study further shows that the same interface can support graph-based Domain-IL with AP-based evaluation, although this portability result is kept separate from HERO-Core because it uses a different modality, loss, and metric.

\paragraph{Limitations.}
HERO-Core currently focuses on image-based FCIL with CIFAR-100 and TinyImageNet. The extended registry and the OGB-MolPCBA case study demonstrate portability, but they are not a dense leaderboard over all modalities and methods. The core protocol also fixes client progression speed and assumes that each client eventually observes every task, which isolates task-order mismatch but does not cover asynchronous client rates or missing-task support. Finally, HERO documents method assumptions and privacy-sensitive artifacts, but it does not replace formal privacy analysis, communication accounting, energy reporting, or deployment-level validation. A full discussion is provided in Appendix~\ref{app:extended_discussion}.

\section{Conclusion}
\label{sec:conclusion}

We introduced \textbf{HERO}, a heterogeneity-aware benchmark library for FCL. HERO unifies stream construction, method execution, metric reporting, and released benchmark files, with HERO-Core providing a dense image-based FCIL benchmark and the portability tier demonstrating extension to graph-based Domain-IL on OGB-MolPCBA. Across the $\alpha$-$\rho$ grid, HERO-Core separates method behavior, exposes bottom-client failures, and shows that client data skew and task-order mismatch affect AFA, AF, and B10 differently. These results motivate FCL evaluation that reports multiple settings and multiple metrics rather than relying on a single global leaderboard. By releasing fixed streams, configurations, method implementations, metrics, and reporting scripts, HERO aims to support cumulative, reproducible, and setting-aware FCL evaluation.

\newpage

{
    \small
    \bibliographystyle{unsrtnat}
    \bibliography{hero}
}


\clearpage

\appendix

\section{Extended Related Work}
\label{app:full_related}

\paragraph{Federated learning benchmarks and libraries.}
Federated learning has motivated several benchmark and software efforts. LEAF provides datasets, metrics, and reference settings for federated evaluation~\citep{caldas2018leaf}. FedScale studies model and system performance at scale and emphasizes realistic client distributions and execution settings~\citep{lai2022fedscale}. Flower provides a flexible framework for federated learning experimentation~\citep{beutel2020flower}. FLamby focuses on realistic cross-silo healthcare federated learning~\citep{ogier2022flamby}. These resources are important, but they mainly target static FL objectives, system-level FL evaluation, or application-specific FL settings. \textbf{HERO} focuses on FCL, where the benchmark must also construct task streams, measure forgetting, and evaluate client-specific task sequences.

Federated optimization under heterogeneity is also closely related. FedProx, SCAFFOLD, FedNova, FedDyn, and Ditto study different ways to address non-IID data, client drift, objective inconsistency, dynamic regularization, and personalization~\citep{li2020fedprox,karimireddy2020scaffold,wang2020tackling,acar2021federated,li2021ditto}. These methods motivate the need to treat client heterogeneity carefully. HERO builds on this view, but moves the evaluation setting from static federated learning to continual streams.

\paragraph{Continual learning and class-incremental evaluation.}
Continual learning studies how models learn from non-stationary data without losing earlier knowledge. Regularization, distillation, replay, and gradient constraints are common families of approaches~\citep{kirkpatrick2017overcoming,li2017learning,lopez2017gradient,chaudhry2018efficient,buzzega2020dark}. Class-incremental learning is especially challenging because the label space grows over time and task identity is not available at inference~\citep{rebuffi2017icarl,hou2019learning,wu2019large,van2022three}. Prior work has also shown that scenario definitions, evaluation metrics, and task assumptions strongly affect conclusions~\citep{hsu2018re,diaz2018metrics,de2021continual,mai2022online}.

Continual learning benchmarks such as CLEAR emphasize realistic streams and long-term non-stationarity~\citep{lin2021clear}. Distribution-shift benchmarks such as WILDS show that evaluation should expose differences between average performance and robustness under shifted conditions~\citep{koh2021wilds}. HERO follows this benchmark philosophy in a federated setting by pairing standard continual-learning metrics with client-aware evaluation.

\paragraph{Federated continual and class-incremental learning.}
FCL combines decentralized training with continual streams. FedWeIT studies weighted inter-client transfer for FCL~\citep{yoon2021federated}. CFeD studies continual federated learning with knowledge distillation~\citep{ma2022continual}. Other work studies broad-network FCL, asynchronous FCL, heterogeneous task streams, online streams, and resource-constrained FCL~\citep{le2021federated,shenaj2023afcl,luopan2023fedknow,wuerkaixi2024accurate,Li2025ResourceConstrainedFC,serra2025federated}.

FCIL has become a common setting for evaluating FCL methods. GLFC introduced a federated class-incremental learning formulation and a class-aware learning strategy~\citep{dong2022federated}. MFCL and related data-free methods study generative replay without storing real past data~\citep{babakniya2023data,qi2023better,nguyen2024overcoming}. TARGET uses exemplar-free distillation~\citep{zhang2023target}. LANDER uses text-enhanced data-free learning~\citep{tran2024text}. Re-Fed and Re-Fed+ study efficient replay strategies~\citep{li2024towards,li2025re}. TagFed studies traceable FCL~\citep{wang2024traceable}. LGA studies real-world federated class-incremental learning~\citep{dong2023no}. FedCBDR studies class-wise balancing replay~\citep{qi2025class}. Prompting, LoRA-based adaptation, gradient correction, and personalization have also been explored~\citep{guo2024pilora,piao2024federated,salami2025closed,zhang2025pfedmxf,zhang2025fedagc}. The diversity of these methods makes unified evaluation important, because conclusions can depend on stream construction, heterogeneity setting, and metric choice.

\paragraph{Regime-aware benchmark evaluation.}
A benchmark is useful when it measures the intended capability and makes the source of difficulty inspectable. In FCL, task split, client data split, client task sequence, resource budget, and method assumptions can vary at the same time. If these factors are not separated, a single result table may hide whether a method is robust to client data skew, task-order mismatch, domain shift, or only one favorable configuration. HERO therefore separates stream construction from method execution and reports the same core metrics across multiple heterogeneity settings.

\section{Dataset Registry and Library Tiers}
\label{app:dataset_registry}

This appendix documents the full dataset and scenario registry included in \textbf{HERO}. HERO-Core is the dense benchmark used in the main paper. HERO-Portability contains extension studies that demonstrate how the same interface can support additional scenarios and modalities. HERO-Extended contains dataset constructors and scenario builders included in the repository for future experiments. Unless explicitly stated, portability and extended datasets are not merged into the HERO-Core leaderboard.

\begin{table}[t]
\centering
\caption{Tiered organization of \textbf{HERO}.}
\label{app:tab:hero_tiers}
\renewcommand\arraystretch{1.12}
\footnotesize
\begin{tabular}{l p{0.30\linewidth} p{0.40\linewidth}}
\toprule
Tier & Role & Examples \\
\midrule
HERO-Core & Dense and comparable evaluation & CIFAR-100, TinyImageNet, image-based FCIL, $\alpha$-$\rho$ grid \\
HERO-Portability & Extension studies outside the main leaderboard & PACS, DomainNet, THUCNews, CORA, OGB-MolPCBA \\
HERO-Extended & Repository registry for future experiments & MNIST variants, CIFAR-10, ImageNet-1K, additional methods and generators \\
\bottomrule
\end{tabular}
\end{table}

\begin{table}[t]
\centering
\renewcommand\arraystretch{1.13}
\caption{
Extended dataset registry in \textbf{HERO}. The core tier is used for dense leaderboard evaluation in the main paper. Portability and extended datasets document the library interface and support future studies, but they are not merged into the core leaderboard unless the same protocol, method set, and metric set are used.
}
\label{app:tab:dataset_registry}
\resizebox{\textwidth}{!}{
\begin{tabular}{llllll}
\toprule
Tier & Dataset & Modality & Scenario & Construction & Main metric \\
\midrule
Core & CIFAR-100~\citep{krizhevsky2009learning} & Image & Class-IL & $10$ tasks of $10$ classes & AFA, AF, B10 \\
Core & TinyImageNet~\citep{le2015tiny} & Image & Class-IL & $10$ tasks of $20$ classes & AFA, AF, B10 \\
\midrule
Portability & PACS~\citep{li2017deeper} & Image & Domain-IL & Domain split & Accuracy, forgetting \\
Portability & DomainNet~\citep{peng2019moment} & Image & Domain-IL & Domain split & Accuracy, forgetting \\
Portability & THUCNews & Text & Class-IL & Category split & Accuracy, forgetting \\
Portability & CORA~\citep{sen2008collective} & Graph & Class-IL & Node-label split & Accuracy, forgetting \\
Portability & OGB-MolPCBA~\citep{hu2020open} & Graph & Domain-IL & Scaffold-grouped domains & Average precision \\
\midrule
Extended & MNIST & Image & Class-IL & Class split & Accuracy, forgetting \\
Extended & RotatedMNIST & Image & Domain-IL & Rotation shift & Accuracy, forgetting \\
Extended & ColoredMNIST & Image & Domain-IL & Color shift & Accuracy, forgetting \\
Extended & CIFAR-10~\citep{krizhevsky2009learning} & Image & Class-IL & Short class-incremental stream & Accuracy, forgetting \\
Extended & ImageNet-1K~\citep{deng2009imagenet} & Image & Class-IL & Long-horizon class-incremental stream & Accuracy, forgetting \\
\bottomrule
\end{tabular}}
\end{table}

\paragraph{Class-incremental learning.}
In Class-IL, the label space grows over time. Each task introduces a disjoint set of classes, and the model predicts over all exposed classes without task identity. HERO-Core uses this setting for CIFAR-100 and TinyImageNet.

\paragraph{Domain-incremental learning.}
In Domain-IL, the output space remains fixed, but the input distribution changes across tasks. Each task corresponds to a domain or distribution group. This setting is used in portability studies such as PACS, DomainNet, and scaffold-grouped OGB-MolPCBA.

\paragraph{Heterogeneous-order FCIL.}
In heterogeneous-order FCIL, clients share the same task split but may encounter tasks in different local sequences. This setting is controlled by $\rho$ in HERO-Core.

\section{Stream Generator Details}
\label{app:generator_details}

This appendix gives additional details for the stream generators used in \textbf{HERO}. The main paper gives the high-level construction. Here, we describe task construction, client data splitting, task-order generation, participation schedules, and construction checks more explicitly.

\subsection{Class-IL task construction}

For a Class-IL dataset with label set $\mathcal{Y}$ and $T$ tasks, the benchmark seed $\xi$ determines a class permutation. The permuted labels are split into disjoint task-level class sets,
\begin{equation}
\mathcal{Y}
=
\bigcup_{t=1}^{T}\mathcal{C}_t,
\qquad
\mathcal{C}_i \cap \mathcal{C}_j = \emptyset
\quad \text{for } i \neq j.
\label{app:eq:class_il_split}
\end{equation}
The same class-to-task split is used for all methods under the same seed.

\subsection{Domain-IL task construction}

For Domain-IL, the output space remains fixed and tasks correspond to domains. Let $d_i$ be the domain label of sample $x_i$. A domain constructor assigns samples to task datasets by
\begin{equation}
\mathcal{D}_t
=
\{(x_i,y_i) \mid d_i=t\}.
\label{app:eq:domain_constructor}
\end{equation}
This construction is used for domain-shift datasets such as PACS and DomainNet. For datasets without native domain labels, HERO can construct domains through controlled transformations or grouping functions, such as rotations, color shifts, or scaffold groups.

\subsection{Task-wise client data split}

For each task $t$ and class $c \in \mathcal{C}_t$, HERO samples
\begin{equation}
\mathbf{p}_{t,c}
\sim
\mathrm{Dir}_K(\alpha\mathbf{1}),
\label{app:eq:dirichlet}
\end{equation}
then assigns samples of class $c$ to clients according to $\mathbf{p}_{t,c}$. The allocation is performed task-wise because the goal is to control within-task client skew without accidentally removing entire tasks from some clients.

\paragraph{Minimum-support repair.}
When $\alpha$ is small, integer rounding can create empty client-task shards. HERO-Core applies a minimum-support repair rule. If a client $k$ has no samples for task $t$, the generator moves a small number of samples from overrepresented clients to client $k$ while preserving the task identity. The repair is deterministic under the benchmark seed and is recorded in the audit log. This repair does not remove statistical heterogeneity, but it prevents task-order experiments from being confounded by missing-task support.

\subsection{Client task sequence generation}

The synchronized reference order is
\begin{equation}
\pi^\star=(1,2,\ldots,T).
\label{app:eq:reference_order}
\end{equation}
For $\rho=0$, every client follows $\pi^\star$. For $\rho>0$, HERO samples client-specific task sequences centered around $\pi^\star$. The generator accepts a candidate order when its normalized Kendall distance to $\pi^\star$ falls within a target interval,
\begin{equation}
\mathcal{A}_{\rho}
=
\left\{
\pi
\mid
\ell_{\rho}
\le
\frac{d_{\tau}(\pi,\pi^\star)}{\binom{T}{2}}
\le
u_{\rho}
\right\}.
\label{app:eq:rho_acceptance}
\end{equation}
Here, $d_{\tau}$ is Kendall distance, and $[\ell_{\rho},u_{\rho}]$ is the acceptance interval for the requested perturbation level. The nominal value $\rho$ controls generation, while the realized order checks describe the actual generated stream.

\subsection{Construction checks}

For client data skew, let $q_{k,t}$ be the empirical class distribution of client $k$ within task $t$. HERO reports
\begin{equation}
\hat{h}_{\mathrm{stat}}
=
\frac{1}{T}
\sum_{t=1}^{T}
\frac{2}{K(K-1)}
\sum_{1 \le i < j \le K}
\mathrm{JS}(q_{i,t}\,\|\,q_{j,t}),
\label{app:eq:h_stat}
\end{equation}
where $\mathrm{JS}$ is Jensen-Shannon divergence~\citep{lin2002divergence}.

For task-order mismatch, HERO reports pairwise client disagreement,
\begin{equation}
\hat{\psi}_{\mathrm{pair}}
=
\frac{2}{K(K-1)}
\sum_{1 \le i < j \le K}
\frac{d_{\tau}(\pi_i,\pi_j)}{\binom{T}{2}},
\label{app:eq:psi_pair}
\end{equation}
and deviation from the synchronized reference order,
\begin{equation}
\hat{\psi}_{\mathrm{ref}}
=
\frac{1}{K}
\sum_{k=1}^{K}
\frac{d_{\tau}(\pi_k,\pi^\star)}{\binom{T}{2}}.
\label{app:eq:psi_ref}
\end{equation}
The first quantity measures disagreement among clients, while the second measures deviation from the canonical synchronized sequence.

\subsection{Participation trace}

For each benchmark stage and communication round, HERO samples a participating client set according to the participation fraction $C$. The resulting participation trace is stored and reused by all methods. This prevents differences in client sampling from being mistaken for algorithmic differences.

\subsection{Generator output}

The generator saves a benchmark stream before any method is executed. The core stream contains the task split, client data split, test split, client task sequence, participation trace, and construction checks. Thus, methods do not regenerate their own task splits, partitions, orders, sampling schedules, or evaluation splits.

\section{Metric Definitions and Report Schema}
\label{app:metrics}

This appendix gives the complete metric definitions used by \textbf{HERO}. HERO-Core uses accuracy-based metrics for image-based FCIL. Portability studies may use task-appropriate metrics, such as average precision for OGB-MolPCBA.

\subsection{Client-local exposed label set}

When clients follow different task sequences, there is no single global class prefix shared by all clients at a stage. For client $k$ at stage $s$, the exposed label set is
\begin{equation}
\mathcal{Y}^{\mathrm{seen}}_{k,s}
=
\bigcup_{r=1}^{s}
\mathcal{C}_{\pi_k(r)}.
\label{app:eq:seen_labels}
\end{equation}
Stage-wise evaluation is client-local Class-IL. The model predicts over $\mathcal{Y}^{\mathrm{seen}}_{k,s}$ without task identity. Final evaluation is performed over the full label space.

\subsection{Accuracy matrix}

Let $A_{k,s,t}$ denote the accuracy on client $k$'s test samples from task $t$ after benchmark stage $s$. Let
\begin{equation}
\tau_{k,t}=\pi_k^{-1}(t)
\label{app:eq:arrival_time}
\end{equation}
be the local stage at which client $k$ first encounters task $t$.

\subsection{Final average accuracy}

Final average accuracy (AFA) is
\begin{equation}
\mathrm{AFA}
=
\frac{1}{KT}
\sum_{k=1}^{K}
\sum_{t=1}^{T}
A_{k,T,t}.
\label{app:eq:afa}
\end{equation}
Higher AFA is better.

\subsection{Average forgetting}

Average forgetting (AF) is
\begin{equation}
\mathrm{AF}
=
\frac{1}{KT}
\sum_{k=1}^{K}
\sum_{t=1}^{T}
\left(
\max_{\tau_{k,t}\le s\le T} A_{k,s,t}
-
A_{k,T,t}
\right).
\label{app:eq:af}
\end{equation}
Lower AF is better.

\subsection{Bottom-10\% client accuracy}

Let
\begin{equation}
a_k
=
\frac{1}{T}
\sum_{t=1}^{T}
A_{k,T,t}
\label{app:eq:client_acc}
\end{equation}
be the final average accuracy of client $k$. Let $\mathcal{K}_{10}$ be the set of clients with the lowest final accuracies, with
\begin{equation}
|\mathcal{K}_{10}|
=
\max(1,\lfloor0.1K\rfloor).
\label{app:eq:k10}
\end{equation}
Bottom-10\% client accuracy (B10) is
\begin{equation}
\mathrm{B10}
=
\frac{1}{|\mathcal{K}_{10}|}
\sum_{k\in\mathcal{K}_{10}} a_k.
\label{app:eq:b10}
\end{equation}
Higher B10 is better.

\subsection{Average precision for OGB-MolPCBA}

For OGB-MolPCBA, HERO reports average precision (AP), which is more appropriate than accuracy for imbalanced multi-label molecular property prediction. Each molecule has a target vector
\begin{equation}
y_i \in \{0,1,\mathrm{NaN}\}^{128}.
\label{app:eq:molpcba_label}
\end{equation}
Missing labels are ignored through an observation mask. The AP for assay $j$ is computed from the precision-recall curve over valid labels, and the final score is the mean AP over assays,
\begin{equation}
\mathrm{AP}
=
\frac{1}{128}
\sum_{j=1}^{128}
\mathrm{AP}_j.
\label{app:eq:molpcba_ap}
\end{equation}

\subsection{Report schema}

The report generator stores one row for each method, dataset, seed, $\alpha$, and $\rho$ cell,
\begin{equation}
\mathrm{report}
=
\left(
\mathrm{dataset},
\mathrm{scenario},
\mathrm{method},
\mathrm{seed},
\alpha,
\rho,
\mathrm{AFA},
\mathrm{AF},
\mathrm{B10},
\hat{h}_{\mathrm{stat}},
\hat{\psi}_{\mathrm{pair}},
\hat{\psi}_{\mathrm{ref}}
\right).
\label{app:eq:report_schema}
\end{equation}
Additional fields store runtime, memory use, method-specific assumptions, and task-specific portability metrics when available.

\section{Artifact Schema and Release Format}
\label{app:artifact_schema}

A benchmark library should expose reusable files so that future methods can be evaluated without regenerating hidden benchmark choices. HERO therefore separates stream generation from method execution. For a fixed benchmark cell, all methods read the same task split, client shards, client task sequences, participation trace, and evaluation split.

\begin{table}[t]
\centering
\caption{Artifact schema in \textbf{HERO}. Each file is generated from a configuration and seed. All methods use the same files for a fixed benchmark cell.}
\label{app:tab:artifact_schema}
\renewcommand\arraystretch{1.12}
\footnotesize
\begin{tabular}{l p{0.70\linewidth}}
\toprule
File & Content \\
\midrule
\texttt{config.yaml} & Dataset, scenario, seed, model settings, optimizer settings, and heterogeneity settings \\
\texttt{tasks.json} & Task split, class sets, domain sets, task indices, and task metadata \\
\texttt{client\_shards.json} & Client-level training and test shards for each task \\
\texttt{orders.json} & Client task sequences, synchronized reference order, and order-generation metadata \\
\texttt{participation.json} & Participating clients for each communication round and stage \\
\texttt{evaluation.json} & Evaluation label sets, evaluation split, and metric settings \\
\texttt{audit.json} & Realized client-skew check, order-mismatch checks, and generator repair logs \\
\texttt{method.yaml} & Method-specific modules, replay budget, generator usage, personalization, pretraining, and tuning settings \\
\texttt{results.csv} & Raw method outputs, seed-level metrics, benchmark-cell identifiers, and runtime metadata \\
\texttt{report.json} & Aggregated metrics, full-grid summaries, setting-wise summaries, and diagnostic statistics \\
\bottomrule
\end{tabular}
\end{table}

\subsection{Recommended repository layout}

\begin{verbatim}
hero/
  configs/
    core/
    portability/
    extended/
  data/
    metadata/
    generated_streams/
  hero/
    datasets/
    generators/
    methods/
    metrics/
    reporting/
  scripts/
    generate_stream.py
    run_method.py
    aggregate_results.py
    make_tables.py
    make_figures.py
  reports/
    core/
    appendix/
\end{verbatim}

\subsection{Versioning}

HERO maintains separate versions for stream construction and method implementation. A change to task splits, client data splits, task-order samplers, evaluation splits, or metrics creates a new benchmark version. A change to method code, hyperparameters, replay budget, or generator implementation creates a new method version. This separation allows users to know whether two results are directly comparable.

\begin{table}[t]
\centering
\caption{Released files in \textbf{HERO}. The core release supports reproduction of the main results. The extended release supports portability studies and future extensions.}
\label{app:tab:release_files}
\renewcommand\arraystretch{1.12}
\footnotesize
\begin{tabular}{l p{0.50\linewidth} l}
\toprule
File group & Content & Tier \\
\midrule
Stream generator & Task split, client data split, and task-sequence generation & Core, Extended \\
Fixed streams & Saved task splits, client data splits, client task sequences, and sampling schedules & Core \\
Dataset registry & Core, portability, and extended dataset metadata & Core, Extended \\
Method collection & Reference baselines and implemented FCL methods & Core, Extended \\
Metric scripts & AFA, AF, B10, AP, and construction checks & Core, Extended \\
Report scripts & Tables, full-grid summaries, and setting-wise reports & Core \\
Configuration files & YAML files for HERO-Core settings and method runs & Core \\
Portability configs & Configuration files for Domain-IL, text, and graph studies & Extended \\
\bottomrule
\end{tabular}
\end{table}

\section{Method Zoo and Implementation Details}
\label{app:method_zoo}

This appendix documents the method zoo implemented or supported by \textbf{HERO}. The table separates methods used in the dense HERO-Core leaderboard from extended methods. Extended methods are important for library coverage, but they are not always included in the main comparable pool because they may require different assumptions, such as personalization, pretrained modules, generators, or resource-specific settings.

\begin{table}[t]
\centering
\renewcommand\arraystretch{1.12}
\caption{
Method coverage in \textbf{HERO}. Core methods are used in the dense HERO-Core leaderboard. Extended methods are implemented or documented in the library and are evaluated in separate tracks when their assumptions differ from the core protocol.
}
\label{app:tab:method_zoo}
\resizebox{\textwidth}{!}{
\begin{tabular}{lllp{3.3cm}p{3.8cm}l}
\toprule
Tier & Family & Method & Native scenario & Main mechanism & Extra assumptions \\
\midrule
Core & Reference FL & Local-Only & Class-IL, Domain-IL & Independent client training & No federation \\
Core & Reference FL & FedAvg~\citep{mcmahan2017communication} & Class-IL, Domain-IL & Server model averaging & No FCIL-specific module \\
Extended & Reference FL & FedProx~\citep{li2020fedprox} & Class-IL, Domain-IL & Proximal regularization & No replay, no generator \\
Extended & Personalized FL & FedALA~\citep{zhang2023fedala} & Class-IL, Domain-IL & Adaptive local aggregation & Personalization \\
Extended & Personalized FL & FedAS~\citep{yang2024fedas} & Class-IL, Domain-IL & Adaptive aggregation & Personalization \\
Extended & Personalized FL & FedL2P~\citep{lee2023fedl2p} & Class-IL, Domain-IL & Federated personalization & Personalization \\
\midrule
Extended & FDG & FedDBE~\citep{zhang2023eliminating} & Domain-IL, distribution shift & Representation-space debiasing & Domain-generalization objective \\
Extended & FDG & FedOMG~\citep{nguyen2025federated} & Domain-IL, distribution shift & On-server matching gradient & Server-side gradient matching \\
\midrule
Core & FCIL & GLFC~\citep{dong2022federated} & Class-IL & Class-aware loss and distillation & FCIL-specific objective \\
Core & FCIL & MFCL~\citep{babakniya2023data} & Class-IL & Data-free replay & Generator, synthetic replay \\
Core & FCIL & TARGET~\citep{zhang2023target} & Class-IL & Exemplar-free distillation & Distillation, auxiliary generation \\
Core & FCIL & LANDER~\citep{tran2024text} & Class-IL & Text-enhanced data-free replay & Semantic side information \\
Core & FCIL & Re-Fed+~\citep{li2025re} & Class-IL, incremental FL & Replay strategy & Replay buffer \\
Core & FCIL & TagFed~\citep{wang2024traceable} & Class-IL & Traceable task-aware learning & Task tracing, distillation \\
Core & FCIL & LGA~\citep{dong2023no} & Class-IL & Real-world FCIL adaptation & Category-aware objective \\
Core & FCIL & FedCBDR~\citep{qi2025class} & Class-IL & Class-wise balancing replay & Replay, class balancing \\
Extended & FCIL & FedCIL~\citep{qi2023better} & Class-IL & Better generative replay & Generator, synthetic replay \\
Extended & FCIL & FedSSI~\citep{li2025fedssi} & Class-IL & Synaptic-intelligence regularization & Rehearsal-free regularization \\
Extended & FCIL & FOT~\citep{bakman2024federated} & Class-IL, Domain-IL & Orthogonal training & Gradient projection \\
Extended & FCIL & PiLoRA~\citep{guo2024pilora} & Class-IL & Prototype-guided LoRA adaptation & PEFT, pretrained modules \\
Extended & FCIL & pFedMxF~\citep{zhang2025pfedmxf} & Class-IL & Personalized frequency aggregation & Personalization, pretrained modules \\
Extended & Online FCL & O-FCL~\citep{serra2025federated} & Online Class-IL, Domain-IL & Uncertainty-aware memory management & Online stream, memory budget \\
\midrule
Extended & Heterogeneous FCL & FedWeIT~\citep{yoon2021federated} & Heterogeneous task streams & Parameter decomposition and transfer & Personalization, parameter expansion \\
Extended & Heterogeneous FCL & AF-FCL~\citep{wuerkaixi2024accurate} & Heterogeneous Class-IL & Accurate forgetting & Forgetting-aware objective \\
Extended & Heterogeneous FCL & STAMP~\citep{nguyen2025spatio} & Heterogeneous Class-IL & Spatio-temporal gradient matching & Gradient alignment \\
Extended & Heterogeneous FCL & FedAGC~\citep{zhang2025fedagc} & Heterogeneous Class-IL & Asymmetric gradient correction & Gradient correction \\
Extended & Heterogeneous FCL & FedTA~\citep{yu2025handling} & Heterogeneous Class-IL & Tail-anchor adaptation & Pretrained model, prototype guidance \\
Extended & Heterogeneous FCL & CAN~\citep{rong2025can} & Heterogeneous Class-IL & Client-navigator generative replay & Generator, client navigation \\
\bottomrule
\end{tabular}}
\end{table}

\subsection{Per-method documentation format}

Each method implementation is documented using the following template.
\begin{enumerate}[leftmargin=*]
\item Core idea.
\item Mechanism for continual learning, transfer, or heterogeneity.
\item Adaptation to the HERO protocol.
\item Method-specific assumptions, such as replay, generation, personalization, or pretraining.
\item Hyperparameters and tuning budget.
\item Code path in the repository.
\end{enumerate}

\section{Benchmark Sanity Checks}
\label{app:sanity_checks}

A benchmark should be discriminative, stable across seeds, and difficult enough to expose meaningful failures. Table~\ref{app:tab:sanity_checks} reports four checks for HERO-Core. The method-separation gap measures whether the benchmark distinguishes methods. The easy-to-joint drop measures whether joint heterogeneity creates a meaningful stress test. The mean seed standard deviation measures stability. The AFA-B10 gap measures whether bottom-client behavior is visible rather than hidden by average accuracy.

\begin{table}[t]
\centering
\caption{Sanity checks for HERO-Core. Values are reported over five benchmark seeds.}
\label{app:tab:sanity_checks}
\renewcommand\arraystretch{1.10}
\footnotesize
\begin{tabular}{lcc}
\toprule
Check & CIFAR-100 & TinyImageNet \\
\midrule
Method-separation AFA gap & \meanstd{19.39}{1.63} & \meanstd{15.79}{1.41} \\
Easy-to-joint AFA drop & \meanstd{11.70}{3.47} & \meanstd{8.97}{1.98} \\
Mean AFA seed std & \meanstd{2.13}{0.43} & \meanstd{1.77}{0.37} \\
Mean AFA-B10 gap & \meanstd{5.73}{2.36} & \meanstd{5.73}{2.81} \\
\bottomrule
\end{tabular}
\end{table}

The method-separation gaps are larger than the seed variation, the easy-to-joint drops show that joint heterogeneity is a meaningful stress test, and the AFA-B10 gaps show that bottom-client performance is a distinct signal.

\section{Training Details and Hyperparameters}
\label{app:training_details}

This appendix summarizes the default HERO-Core training protocol. The exact values should match the released \texttt{config.yaml} files. When a method requires additional modules, such as a generator, replay buffer, or distillation model, those settings are stored in the method-specific configuration file.

\begin{table}[t]
\centering
\caption{Default training settings for HERO-Core.}
\label{app:tab:training_details}
\renewcommand\arraystretch{1.12}
\footnotesize
\begin{tabular}{l p{0.60\linewidth}}
\toprule
Component & Setting \\
\midrule
Datasets & CIFAR-100, TinyImageNet \\
Scenario & Image-based Class-IL \\
Clients & $K=20$ \\
Tasks & $T=10$ \\
Participation & $C=0.25$ clients per communication round \\
Local training & $E=5$ local epochs per participating client \\
Communication & $R=40$ rounds per stage \\
Seeds & $5$ benchmark seeds \\
Evaluation & Client-local Class-IL during stages, full label space at final evaluation \\
Reported metrics & AFA, AF, B10 \\
Method assumptions & Replay, generation, distillation, personalization, and pretraining are recorded in \texttt{method.yaml} \\
\bottomrule
\end{tabular}
\end{table}

\paragraph{Fairness rules.}
All methods use the same generated stream for a fixed dataset, seed, $\alpha$, and $\rho$. They also use the same participation trace. Methods cannot access future tasks, future client task sequences, future data, or test data. Method-specific modules are retained when they are part of the original design, but their assumptions are reported in the method collection.

\paragraph{Hyperparameter reporting.}
For each method, the released configuration records optimizer type, learning rate, weight decay, batch size, data augmentation, replay budget, generator settings, distillation temperature, and personalization settings when applicable. If a method uses a published default hyperparameter, the configuration records the source. If a hyperparameter is tuned, the tuning range and selected value are stored.

\section{Compute Resources}
\label{app:compute_resources}

This appendix reports the compute resources used to run the experiments in the main paper. The goal is to make the benchmark cost transparent and to help users plan reproduction runs. All reported experiments were run with one GPU per training job. We used NVIDIA RTX A5000 GPUs with 24GB GPU memory. Unless otherwise stated, each job used one GPU, CPU data-loading workers, and local SSD storage for dataset shards and generated benchmark streams.

\paragraph{Job accounting.}
For HERO-Core, one training job corresponds to one method, one dataset, one benchmark seed, and one $\alpha$-$\rho$ cell. HERO-Core contains $2$ datasets, $10$ methods, $5$ seeds, and $3\times3$ heterogeneity cells. Thus, reproducing the full HERO-Core grid requires
\[
2 \times 10 \times 5 \times 9 = 900
\]
training jobs. The OGB-MolPCBA portability study uses $3$ methods, $5$ seeds, and $3$ scaffold-domain granularities, giving
\[
3 \times 5 \times 3 = 45
\]
training jobs. These jobs are independent and can be run in parallel across multiple GPUs.

\paragraph{Wall-clock time and memory.}
Table~\ref{app:tab:compute_resources} reports the approximate compute envelope for the reported experiments. Runtime depends on implementation details, data-loading speed, and method-specific modules such as replay buffers, generators, or distillation models. We therefore report ranges rather than a single number. The estimates count only the reported runs needed to reproduce the main tables and figures, and do not include preliminary debugging or failed exploratory runs.

\begin{table}[t]
\centering
\caption{
Compute resources for reproducing the reported experiments. Each training job uses one NVIDIA RTX A5000 GPU with 24GB GPU memory. Wall-clock time is reported per method-seed-benchmark-cell job, and total GPU-hours are estimated from the number of reported jobs. Exact runtime may vary with data-loading speed and method-specific modules.
}
\label{app:tab:compute_resources}
\renewcommand\arraystretch{1.12}
\footnotesize
\setlength{\tabcolsep}{4pt}
\begin{tabular}{l c c c c}
\toprule
Experiment group & Jobs & GPU per job & Peak GPU memory & Wall-clock per job \\
\midrule
CIFAR-100 HERO-Core & $10 \times 5 \times 9 = 450$ & 1$\times$ A5000 & $8$-$16$GB & $2.5$-$5.0$ hours \\
TinyImageNet HERO-Core & $10 \times 5 \times 9 = 450$ & 1$\times$ A5000 & $12$-$20$GB & $4.0$-$8.0$ hours \\
OGB-MolPCBA portability & $3 \times 5 \times 3 = 45$ & 1$\times$ A5000 & $10$-$18$GB & $1.0$-$3.0$ hours \\
\bottomrule
\end{tabular}
\end{table}

\paragraph{Total compute estimate.}
Using the ranges in Table~\ref{app:tab:compute_resources}, reproducing the full HERO-Core grid requires approximately $2{,}925$-$5{,}850$ A5000 GPU-hours. The OGB-MolPCBA portability study requires approximately $45$-$135$ additional A5000 GPU-hours. Since all jobs are independent across methods, seeds, datasets, and heterogeneity cells, the wall-clock reproduction time can be reduced by distributing jobs over multiple A5000 GPUs.

\paragraph{Storage and CPU requirements.}
The experiments do not require multi-GPU synchronization within a single job. We recommend at least $64$GB host RAM and local SSD storage for generated benchmark streams, cached datasets, and result logs. CPU requirements are mainly for data loading and preprocessing; the training scripts use standard CPU data-loader workers and do not rely on specialized CPU hardware.

\paragraph{Reproduction scope.}
The compute estimates above cover the reported benchmark tables and figures. Additional exploratory runs, hyperparameter debugging, or future method additions are not required to reproduce the main results and should be accounted for separately.

\section{Extended Discussion and Limitations}
\label{app:extended_discussion}

\paragraph{What HERO changes about FCL evaluation.}
The main value of \textbf{HERO} is that it turns FCL evaluation from paper-specific protocols into a reusable benchmark library. Instead of treating the dataset split, client split, task sequence, training protocol, and reporting rule as implicit choices inside each paper, HERO stores these choices as explicit benchmark streams and released files. This separation makes results easier to audit. When a method improves, users can check whether the improvement comes from the method itself, the task split, the client data split, the client task sequence, or the metric used to summarize performance.

\paragraph{Why AFA, AF, and B10 should be reported together.}
The core experiments show that one metric is not enough for FCL. Final average accuracy (AFA) measures average retained performance at the end of training. Average forgetting (AF) measures how much previously learned knowledge is lost. Bottom-10\% client accuracy (B10) measures the performance of the lowest-performing clients. These metrics capture different failure modes. For example, FedCBDR is strong in average performance under easier settings, while TagFed and LGA have stronger bottom-client robustness under stronger heterogeneity. Thus, average accuracy can hide failures on difficult clients, and FCL evaluation should report bottom-client behavior together with average performance and forgetting.

\paragraph{Why portability is reported separately.}
The OGB-MolPCBA case study shows that HERO can support graph-based Domain-IL with AP-based evaluation. This study is useful because it tests whether the same stream interface can handle a different modality, a fixed multi-label prediction space, and scaffold-induced domain shifts. However, it is not merged with HERO-Core because it uses graph inputs, a different prediction loss, and AP rather than accuracy. This separation is intentional. It lets HERO demonstrate extensibility without turning incompatible settings into one mixed leaderboard.

\paragraph{Core versus extended coverage.}
The tiered organization is important for honest reporting. HERO-Core is dense and comparable. HERO-Portability demonstrates that the stream and reporting interface extends to other scenarios and modalities. HERO-Extended documents additional constructors and method implementations for future work. These tiers should not be collapsed into one score, because different datasets may require different architectures, losses, metrics, and method assumptions. A method that is meaningful for image-based FCIL may not be directly comparable to a method designed for graph-based Domain-IL or online FCL.

\paragraph{Limitations of the current core benchmark.}
The current HERO-Core leaderboard focuses on image-based FCIL with CIFAR-100 and TinyImageNet. Although these datasets are common in FCIL evaluation and allow controlled comparison across methods, they do not cover all realistic FCL deployments. The core protocol also fixes client progression speed, which isolates task-order mismatch but does not fully model asynchronous client rates. In addition, HERO-Core assumes that each client eventually observes every task. This avoids task-support mismatch, but it does not cover settings where some clients never observe some tasks. Future versions can extend the generator to support missing-task clients, asynchronous local clocks, and more deployment-like participation patterns.

\paragraph{Limitations of portability studies.}
The portability tier demonstrates that HERO can support additional modalities and scenarios, but it is not yet a dense leaderboard over all datasets, all methods, and all metrics. The OGB-MolPCBA case study is designed to test the stream interface under graph-based Domain-IL, not to provide a complete graph FCL benchmark. Similarly, additional text, graph, and domain-incremental datasets in the registry should be interpreted as supported constructors or portability examples unless they are evaluated under a matched method set and a matched protocol.

\paragraph{Responsible use.}
High performance on \textbf{HERO} should not be interpreted as a deployment guarantee. Federated learning can still leak information through model updates, and methods that exchange prototypes, synthetic samples, auxiliary statistics, or generator outputs may introduce additional privacy-sensitive artifacts. HERO documents method assumptions, but it does not replace formal privacy analysis, communication accounting, energy reporting, or system-level safety evaluation. These concerns should be studied before deploying FCL methods in sensitive real-world applications.

\section{Full Single-Axis Anchor Results}
\label{app:anchor_results}

The main paper reports easy, mild, and joint-hard settings. This appendix reports the two single-axis settings. Alpha-hard is $(\alpha=0.10,\rho=0.00)$ and isolates client data skew. Rho-hard is $(\alpha=100.00,\rho=1.00)$ and isolates task-order mismatch.

\begin{table}[t]
\centering
\caption{Single-axis anchor results on CIFAR-100, reported as mean $\pm$ standard deviation over five seeds. Higher is better for AFA and B10. Lower is better for AF.}
\label{app:tab:single_axis_cifar}
\renewcommand\arraystretch{1.12}
\footnotesize
\setlength{\tabcolsep}{3pt}
\begin{tabular}{lccc ccc}
\toprule
\multirow{2}{*}{Method}
& \multicolumn{3}{c}{Alpha-hard}
& \multicolumn{3}{c}{Rho-hard} \\
\cmidrule(lr){2-4}\cmidrule(lr){5-7}
& AFA $\uparrow$ & AF $\downarrow$ & B10 $\uparrow$
& AFA $\uparrow$ & AF $\downarrow$ & B10 $\uparrow$ \\
\midrule
Local-Only & \meanstd{26.51}{2.33} & \meanstd{64.31}{2.37} & \meanstd{17.22}{2.13} & \meanstd{24.84}{2.41} & \meanstd{67.52}{2.43} & \meanstd{22.07}{2.31} \\
FedAvg     & \meanstd{29.44}{2.53} & \meanstd{66.43}{2.43} & \meanstd{19.03}{2.23} & \meanstd{27.62}{2.73} & \meanstd{69.61}{2.51} & \meanstd{23.73}{2.43} \\
GLFC       & \meanstd{41.07}{2.03} & \meanstd{43.13}{1.83} & \meanstd{33.51}{2.13} & \meanstd{39.93}{2.23} & \meanstd{46.12}{1.93} & \meanstd{38.73}{2.03} \\
MFCL       & \meanstd{36.81}{2.03} & \meanstd{41.47}{1.73} & \meanstd{30.72}{2.23} & \meanstd{35.52}{2.23} & \meanstd{44.71}{1.83} & \meanstd{33.82}{2.03} \\
TARGET     & \meanstd{27.32}{1.93} & \meanstd{44.81}{1.83} & \meanstd{23.21}{2.13} & \meanstd{25.73}{2.13} & \meanstd{47.53}{1.93} & \meanstd{24.63}{2.03} \\
LANDER     & \meanstd{35.54}{1.93} & \meanstd{33.23}{1.63} & \meanstd{29.72}{2.13} & \meanstd{34.21}{2.03} & \meanstd{37.23}{1.73} & \meanstd{32.54}{1.93} \\
Re-Fed+    & \meanstd{39.52}{1.93} & \meanstd{36.87}{1.73} & \meanstd{32.44}{2.13} & \meanstd{38.62}{2.03} & \meanstd{40.11}{1.83} & \meanstd{35.73}{2.03} \\
TagFed     & \meanstd{44.13}{1.83} & \meanstd{33.41}{1.53} & \meanstd{39.42}{2.03} & \meanstd{45.51}{1.93} & \meanstd{33.34}{1.53} & \meanstd{42.08}{1.83} \\
LGA        & \meanstd{43.32}{1.83} & \meanstd{33.93}{1.53} & \meanstd{38.64}{2.03} & \meanstd{44.84}{2.03} & \meanstd{34.01}{1.53} & \meanstd{41.61}{1.93} \\
FedCBDR    & \meanstd{45.92}{1.73} & \meanstd{32.33}{1.43} & \meanstd{38.26}{1.93} & \meanstd{41.14}{1.83} & \meanstd{37.31}{1.73} & \meanstd{38.71}{1.83} \\
\bottomrule
\end{tabular}
\end{table}

\begin{table}[t]
\centering
\caption{Single-axis anchor results on TinyImageNet, reported as mean $\pm$ standard deviation over five seeds. Higher is better for AFA and B10. Lower is better for AF.}
\label{app:tab:single_axis_tiny}
\renewcommand\arraystretch{1.12}
\footnotesize
\setlength{\tabcolsep}{3pt}
\begin{tabular}{lccc ccc}
\toprule
\multirow{2}{*}{Method}
& \multicolumn{3}{c}{Alpha-hard}
& \multicolumn{3}{c}{Rho-hard} \\
\cmidrule(lr){2-4}\cmidrule(lr){5-7}
& AFA $\uparrow$ & AF $\downarrow$ & B10 $\uparrow$
& AFA $\uparrow$ & AF $\downarrow$ & B10 $\uparrow$ \\
\midrule
Local-Only & \meanstd{16.93}{1.93} & \meanstd{67.21}{2.23} & \meanstd{8.71}{1.33} & \meanstd{15.83}{2.03} & \meanstd{70.13}{2.33} & \meanstd{11.21}{1.43} \\
FedAvg     & \meanstd{17.62}{2.03} & \meanstd{69.83}{2.33} & \meanstd{8.94}{1.43} & \meanstd{16.23}{2.13} & \meanstd{72.42}{2.43} & \meanstd{11.63}{1.53} \\
GLFC       & \meanstd{27.83}{1.83} & \meanstd{47.91}{1.93} & \meanstd{18.23}{1.43} & \meanstd{26.21}{1.93} & \meanstd{52.33}{2.03} & \meanstd{22.83}{1.53} \\
MFCL       & \meanstd{24.31}{1.83} & \meanstd{45.72}{1.83} & \meanstd{16.91}{1.43} & \meanstd{23.71}{1.93} & \meanstd{49.13}{1.93} & \meanstd{19.41}{1.53} \\
TARGET     & \meanstd{22.52}{1.73} & \meanstd{39.73}{1.73} & \meanstd{16.23}{1.43} & \meanstd{21.82}{1.83} & \meanstd{44.93}{1.83} & \meanstd{18.72}{1.43} \\
LANDER     & \meanstd{25.93}{1.73} & \meanstd{32.83}{1.63} & \meanstd{20.73}{1.43} & \meanstd{24.73}{1.83} & \meanstd{38.32}{1.73} & \meanstd{22.83}{1.43} \\
Re-Fed+    & \meanstd{27.11}{1.73} & \meanstd{36.74}{1.73} & \meanstd{20.41}{1.43} & \meanstd{26.17}{1.83} & \meanstd{42.41}{1.83} & \meanstd{22.31}{1.43} \\
TagFed     & \meanstd{32.21}{1.63} & \meanstd{35.13}{1.53} & \meanstd{27.63}{1.33} & \meanstd{32.61}{1.73} & \meanstd{34.93}{1.53} & \meanstd{29.51}{1.23} \\
LGA        & \meanstd{31.51}{1.63} & \meanstd{36.41}{1.53} & \meanstd{26.72}{1.33} & \meanstd{31.72}{1.73} & \meanstd{36.43}{1.53} & \meanstd{28.42}{1.33} \\
FedCBDR    & \meanstd{31.83}{1.53} & \meanstd{37.92}{1.53} & \meanstd{25.93}{1.23} & \meanstd{30.21}{1.63} & \meanstd{41.82}{1.63} & \meanstd{26.37}{1.33} \\
\bottomrule
\end{tabular}
\end{table}

\section{Axis Sensitivity}
\label{app:axis_sensitivity}

This appendix reports per-method sensitivity on CIFAR-100. Values are changes relative to the easy setting $(\alpha=100.00,\rho=0.00)$. Alpha-hard isolates client data skew, while rho-hard isolates task-order mismatch. Higher values are better for $\Delta$AFA and $\Delta$B10 when the changes are less negative, while lower $\Delta$AF is better.

\begin{table}[t]
\centering
\caption{Per-method sensitivity on CIFAR-100. Values are changes relative to the easy setting $(\alpha=100.00,\rho=0.00)$.}
\label{app:tab:axis_sensitivity}
\renewcommand\arraystretch{1.10}
\footnotesize
\setlength{\tabcolsep}{3pt}
\begin{tabular}{lrrrrrr}
\toprule
\multirow{2}{*}{Method}
& \multicolumn{3}{c}{Alpha-hard}
& \multicolumn{3}{c}{Rho-hard} \\
\cmidrule(lr){2-4}\cmidrule(lr){5-7}
& $\Delta$AFA & $\Delta$AF & $\Delta$B10
& $\Delta$AFA & $\Delta$AF & $\Delta$B10 \\
\midrule
Local-Only & $-7.70$ & $+8.94$  & $-14.61$ & $-9.37$  & $+12.15$ & $-9.76$ \\
FedAvg     & $-7.99$ & $+9.24$  & $-15.49$ & $-9.81$  & $+12.42$ & $-10.79$ \\
GLFC       & $-9.04$ & $+8.86$  & $-14.11$ & $-10.18$ & $+11.85$ & $-8.89$ \\
MFCL       & $-8.02$ & $+8.86$  & $-10.67$ & $-9.31$  & $+12.10$ & $-7.57$ \\
TARGET     & $-5.47$ & $+8.38$  & $-6.91$  & $-7.06$  & $+11.10$ & $-5.49$ \\
LANDER     & $-7.67$ & $+7.60$  & $-11.05$ & $-9.00$  & $+11.60$ & $-8.23$ \\
Re-Fed+    & $-8.37$ & $+8.70$  & $-11.92$ & $-9.27$  & $+11.94$ & $-8.63$ \\
TagFed     & $-6.60$ & $+8.80$  & $-8.52$  & $-5.22$  & $+8.73$  & $-5.86$ \\
LGA        & $-6.35$ & $+8.52$  & $-8.18$  & $-4.83$  & $+8.60$  & $-5.21$ \\
FedCBDR    & $-7.99$ & $+9.60$  & $-12.60$ & $-12.77$ & $+14.58$ & $-12.15$ \\
\bottomrule
\end{tabular}
\end{table}

The table shows that client data skew and task-order mismatch stress methods differently. Alpha-hard settings tend to hurt bottom-client accuracy more strongly, while rho-hard settings tend to increase forgetting and reduce average performance for methods that depend more on synchronized client progress.

\section{Full-Grid Results}
\label{app:full_grid_results}

The main paper reports the full-grid robustness summary. The complete per-cell result table contains $2$ datasets, $9$ heterogeneity cells, $10$ methods, and $3$ metrics, which gives $540$ scalar metric entries before seed-level values. To avoid an excessively long appendix table, HERO stores the complete per-cell results in \texttt{results.csv} and \texttt{report.json}. The released report includes every dataset, method, seed, $\alpha$, $\rho$, AFA, AF, B10, and construction check.

\paragraph{Figures.}
The regime profile figure and full-grid robustness scatter in the main paper are generated from the same report files. The regime profile uses the easy, mild, and joint-hard settings. The robustness scatter uses average AFA over all cells, worst-cell B10, and B10 drop from easy to joint-hard.

\paragraph{Recommended full-grid report fields.}
For reproducibility, each full-grid row should include
\[
(\mathrm{dataset},\mathrm{method},\mathrm{seed},\alpha,\rho,\mathrm{AFA},\mathrm{AF},\mathrm{B10},
\hat{h}_{\mathrm{stat}},\hat{\psi}_{\mathrm{pair}},\hat{\psi}_{\mathrm{ref}}).
\]
This schema is sufficient to regenerate all main tables and figures.

\section{Portability Studies}
\label{app:portability}

This appendix summarizes portability studies outside HERO-Core. These studies are not part of the core leaderboard because they may use different backbones, losses, metrics, or data modalities. Their purpose is to show that the same stream and reporting interface can support broader FCL settings.

\begin{table}[t]
\centering
\caption{Portability studies in the \textbf{HERO} library. These results are not part of the core leaderboard.}
\label{app:tab:portability_summary}
\renewcommand\arraystretch{1.10}
\footnotesize
\begin{tabular}{l l c c c}
\toprule
Dataset & Scenario & Metric & Average score & Forgetting \\
\midrule
PACS & Domain-IL & Accuracy & \meanstd{71.43}{1.87} & \meanstd{6.31}{0.72} \\
DomainNet & Domain-IL & Accuracy & \meanstd{42.87}{2.31} & \meanstd{12.74}{1.43} \\
THUCNews & Class-IL & Accuracy & \meanstd{67.29}{1.92} & \meanstd{9.13}{1.04} \\
CORA & Class-IL & Accuracy & \meanstd{58.46}{2.17} & \meanstd{7.82}{0.91} \\
OGB-MolPCBA & Domain-IL & AP & \meanstd{24.61}{0.83} & \meanstd{3.27}{0.49} \\
\bottomrule
\end{tabular}
\end{table}

\subsection{OGB-MolPCBA}

For OGB-MolPCBA, HERO reports average precision (AP), which is more appropriate than accuracy under highly imbalanced multi-label molecular property prediction~\citep{hu2020open}. Each molecule has a label vector
\[
y_i \in \{0,1,\mathrm{NaN}\}^{128},
\]
where each coordinate corresponds to one biological assay. Missing labels are excluded from the loss and metric computation.

\paragraph{Scaffold-grouped domains.}
We construct a Domain-IL benchmark by inducing domains from molecular scaffolds. Let $s_i$ denote the scaffold ID of molecule $x_i$. Since the original dataset contains many scaffolds, using each scaffold as an individual domain would produce many extremely small domains. We therefore define a grouping function
\[
g:\mathcal{S}\rightarrow\{1,\ldots,K\},
\]
where $\mathcal{S}$ is the set of scaffold IDs and $K$ is the desired number of grouped domains. Each molecule is assigned a domain ID $d_i=g(s_i)$. In the portability experiments, we evaluate $K\in\{5,15,25\}$.

\paragraph{Training with missing labels.}
Let $m_{ij}=1$ if the label for molecule $i$ and assay $j$ is observed, and $m_{ij}=0$ otherwise. HERO optimizes a masked binary cross-entropy objective,
\[
\mathcal{L}
=
\frac{1}{\sum_{i,j}m_{ij}}
\sum_{i,j}
m_{ij}\cdot \ell(y_{ij},\hat{y}_{ij}).
\]
This objective ensures that entries with missing labels do not contribute to the gradient.

\begin{figure}[t]
    \centering
    \begin{subfigure}[t]{0.32\textwidth}
        \centering
        \includegraphics[width=\linewidth]{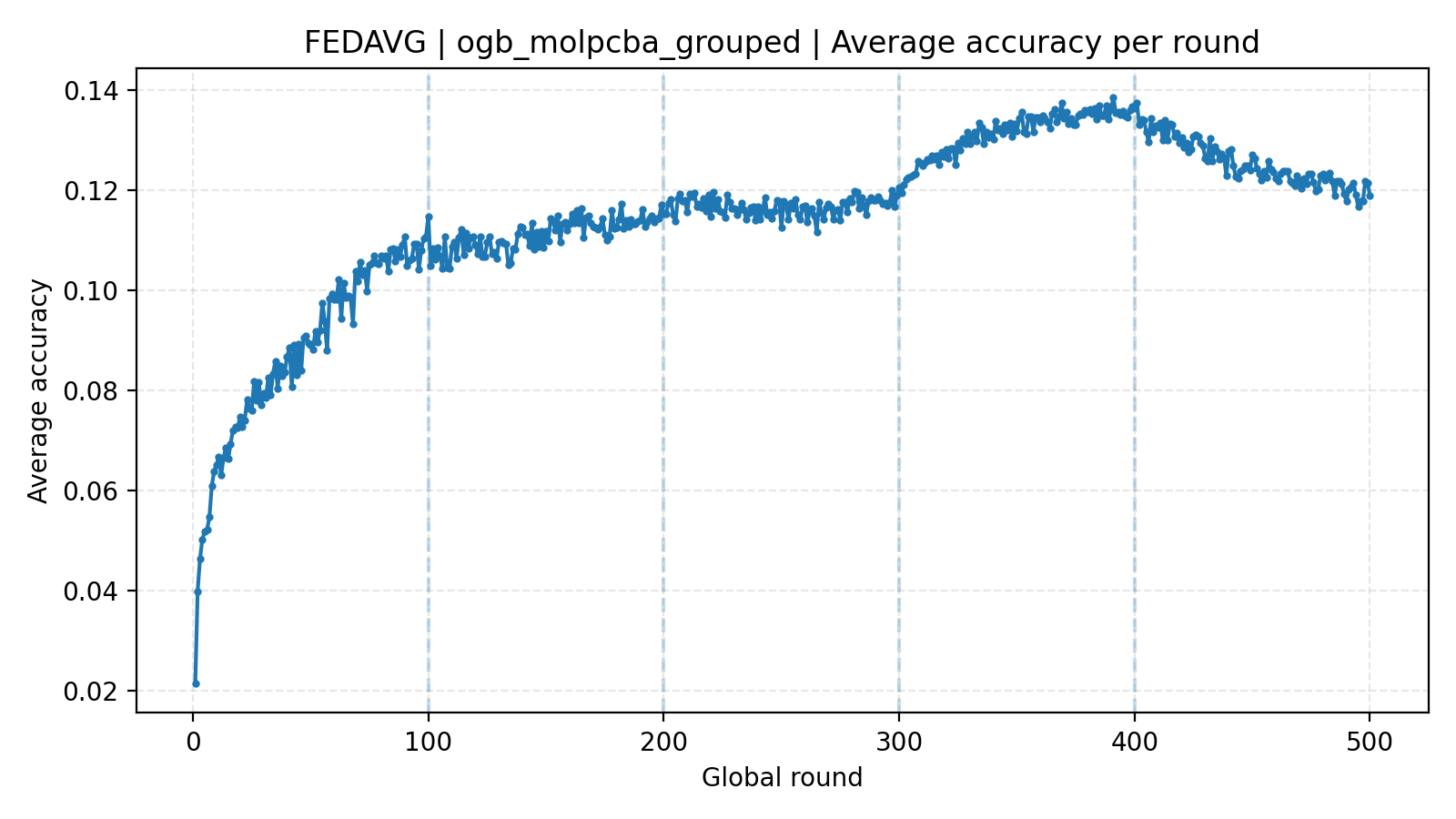}
        \caption{$K=5$ domains}
    \end{subfigure}
    \hfill
    \begin{subfigure}[t]{0.32\textwidth}
        \centering
        \includegraphics[width=\linewidth]{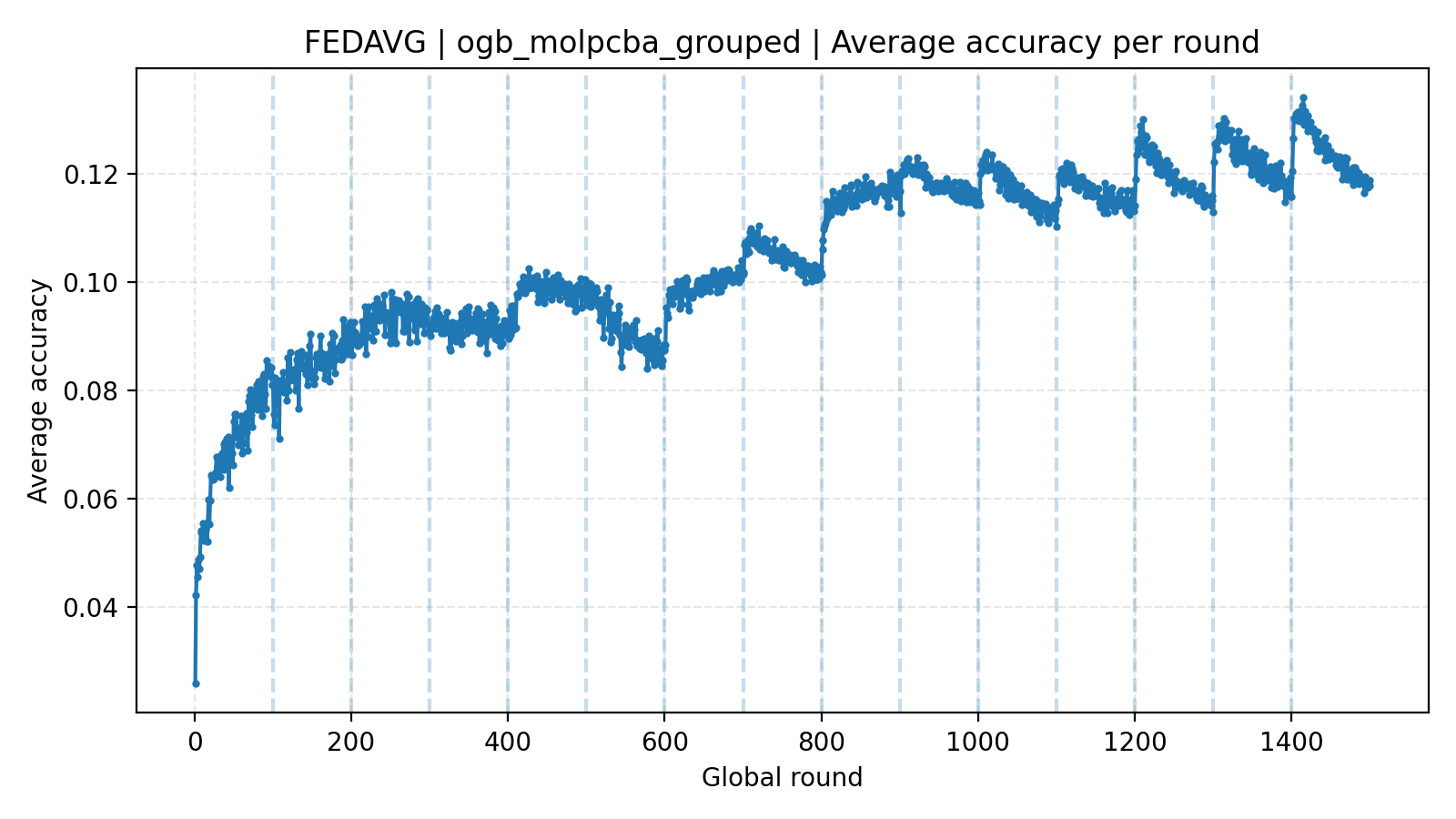}
        \caption{$K=15$ domains}
    \end{subfigure}
    \hfill
    \begin{subfigure}[t]{0.32\textwidth}
        \centering
        \includegraphics[width=\linewidth]{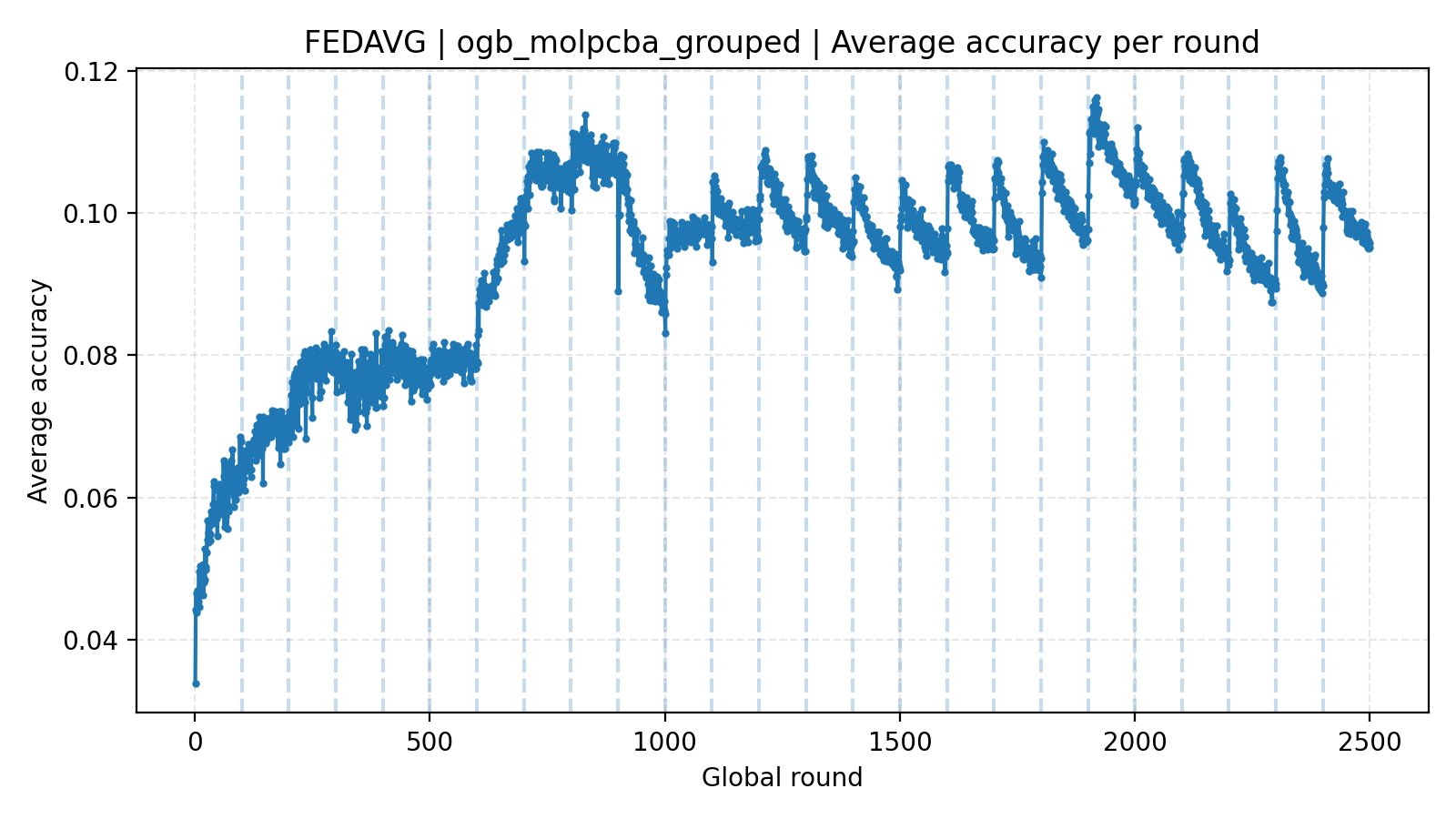}
        \caption{$K=25$ domains}
    \end{subfigure}
    \caption{Average precision on scaffold-grouped OGB-MolPCBA under increasing domain granularity. Finer scaffold partitions induce more frequent and more pronounced fluctuations.}
    \label{fig:molpcba-ap}
\end{figure}

\begin{figure}[t]
    \centering
    \begin{subfigure}[t]{0.32\textwidth}
        \centering
        \includegraphics[width=\linewidth]{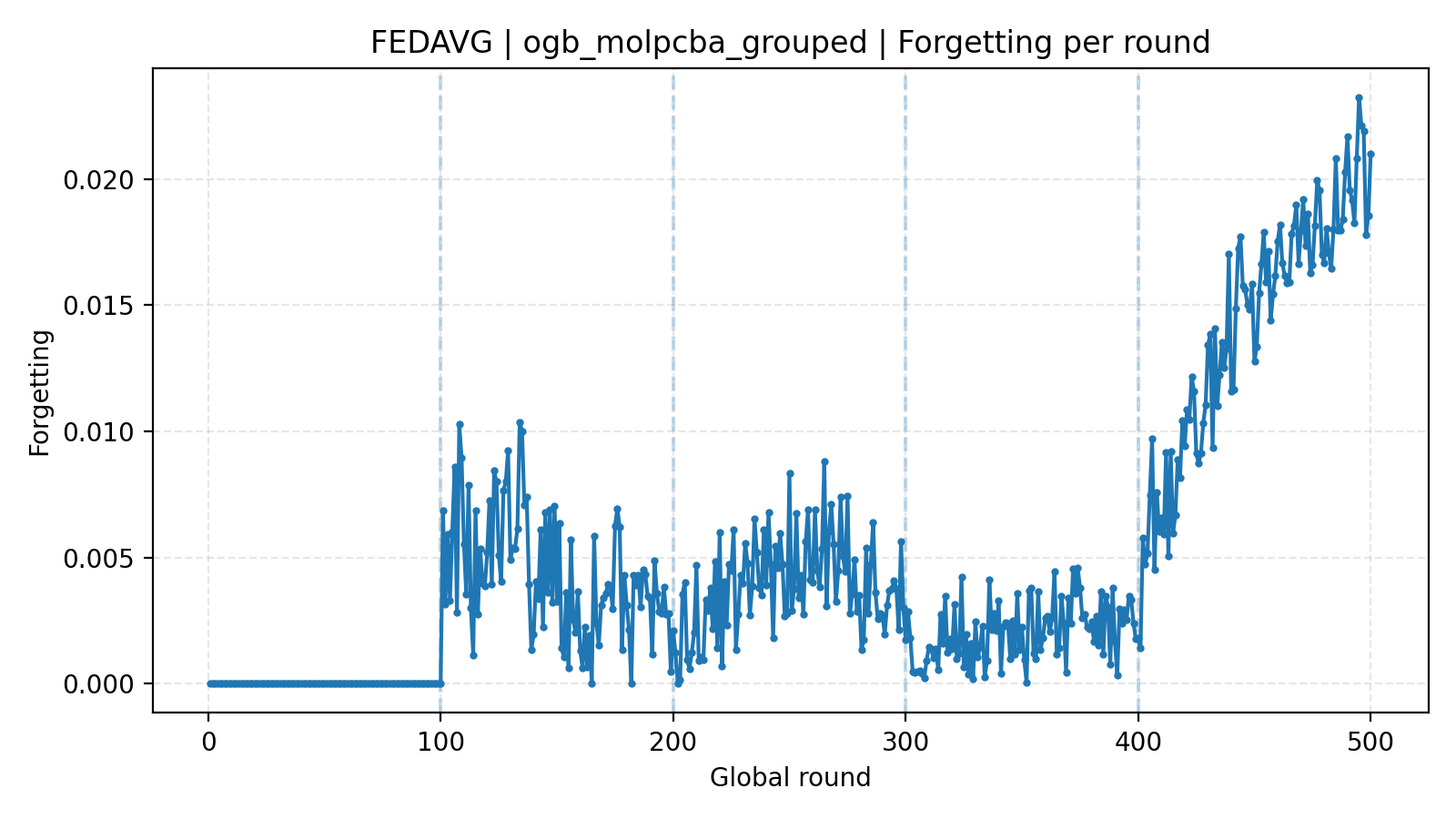}
        \caption{$K=5$ domains}
    \end{subfigure}
    \hfill
    \begin{subfigure}[t]{0.32\textwidth}
        \centering
        \includegraphics[width=\linewidth]{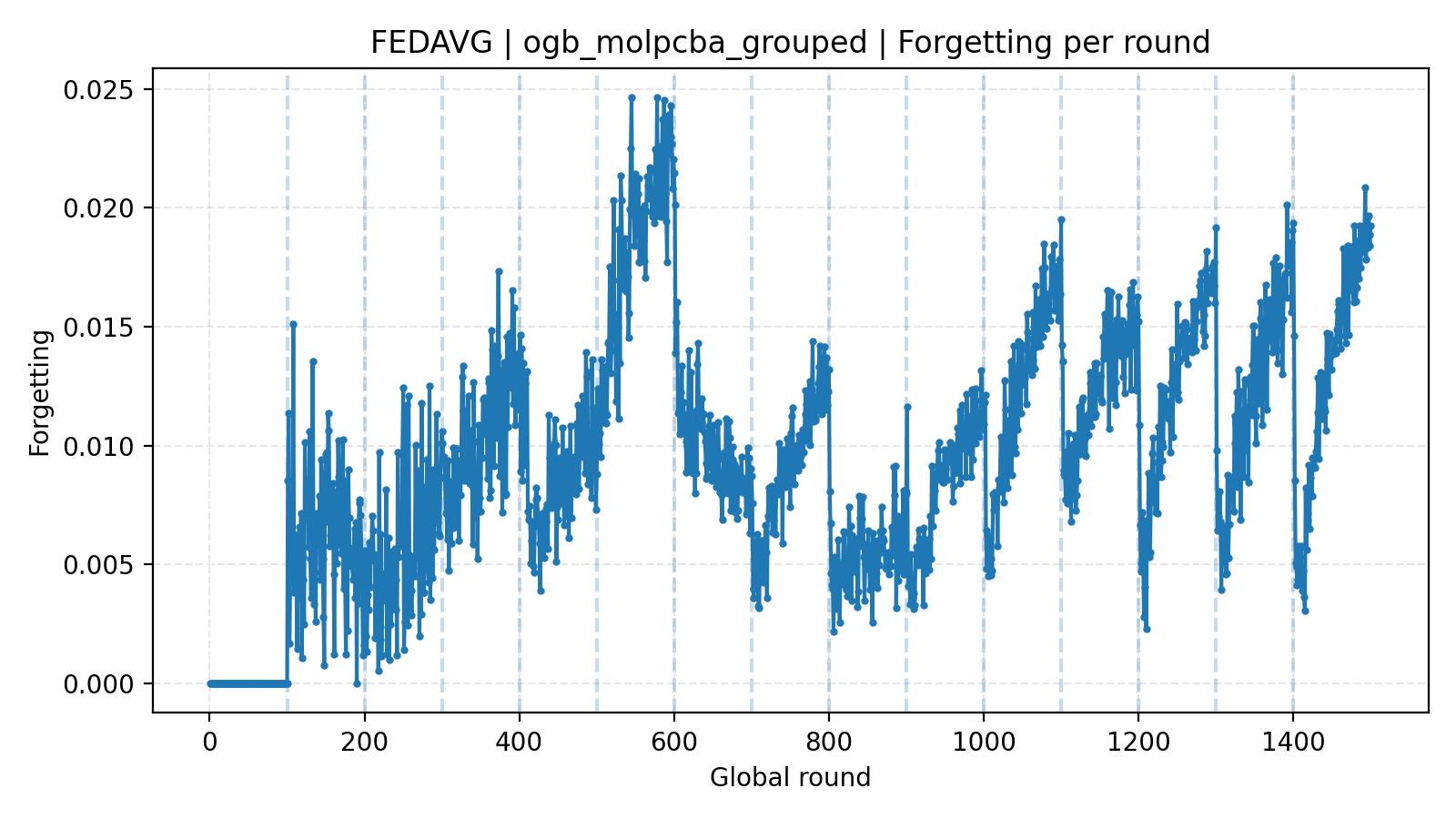}
        \caption{$K=15$ domains}
    \end{subfigure}
    \hfill
    \begin{subfigure}[t]{0.32\textwidth}
        \centering
        \includegraphics[width=\linewidth]{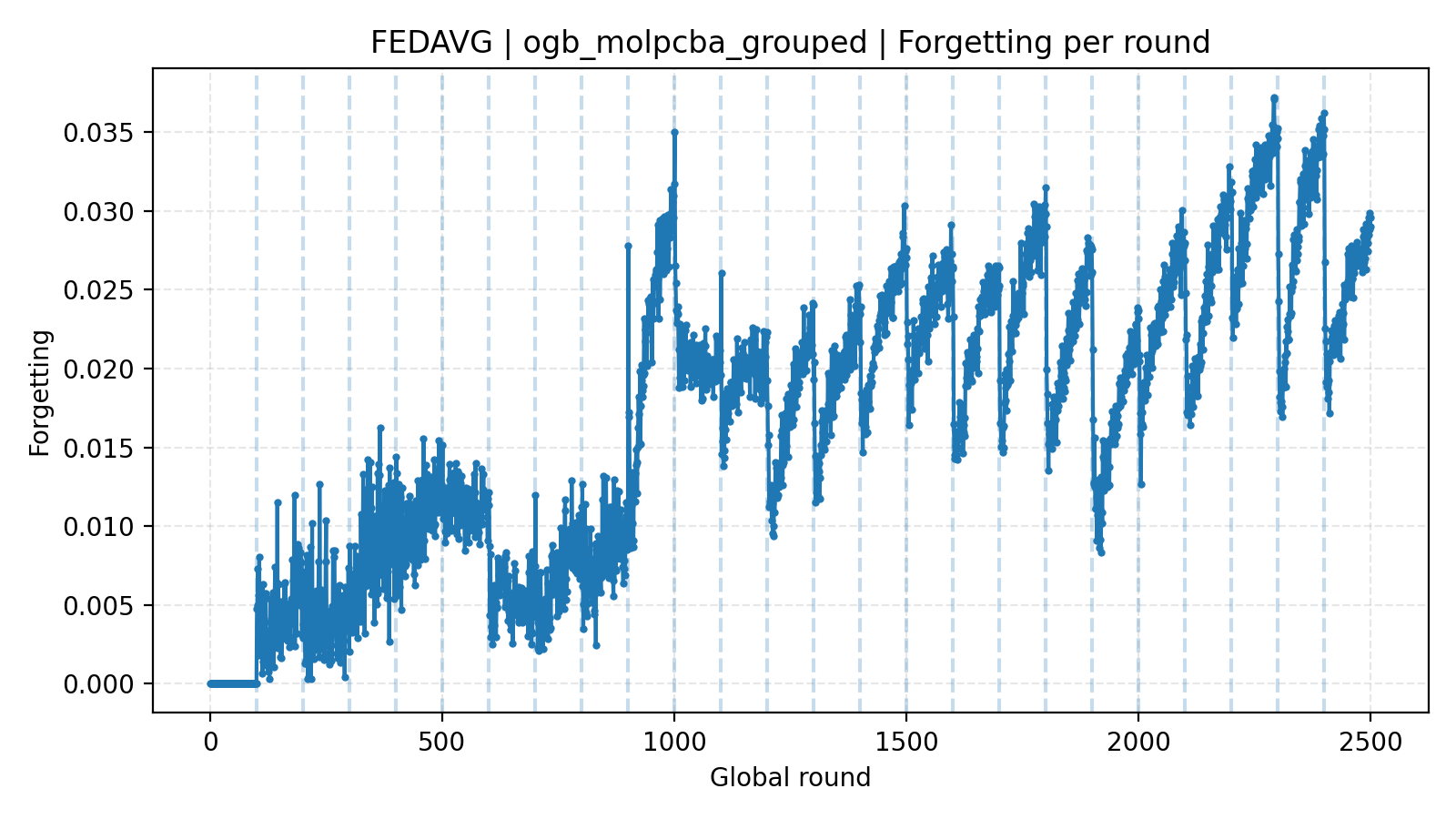}
        \caption{$K=25$ domains}
    \end{subfigure}
    \caption{Forgetting on scaffold-grouped OGB-MolPCBA. Forgetting becomes larger and more unstable as the number of scaffold groups increases.}
    \label{fig:molpcba-forgetting}
\end{figure}

\begin{figure}[t]
    \centering
    \begin{subfigure}[t]{0.32\textwidth}
        \centering
        \includegraphics[width=\linewidth]{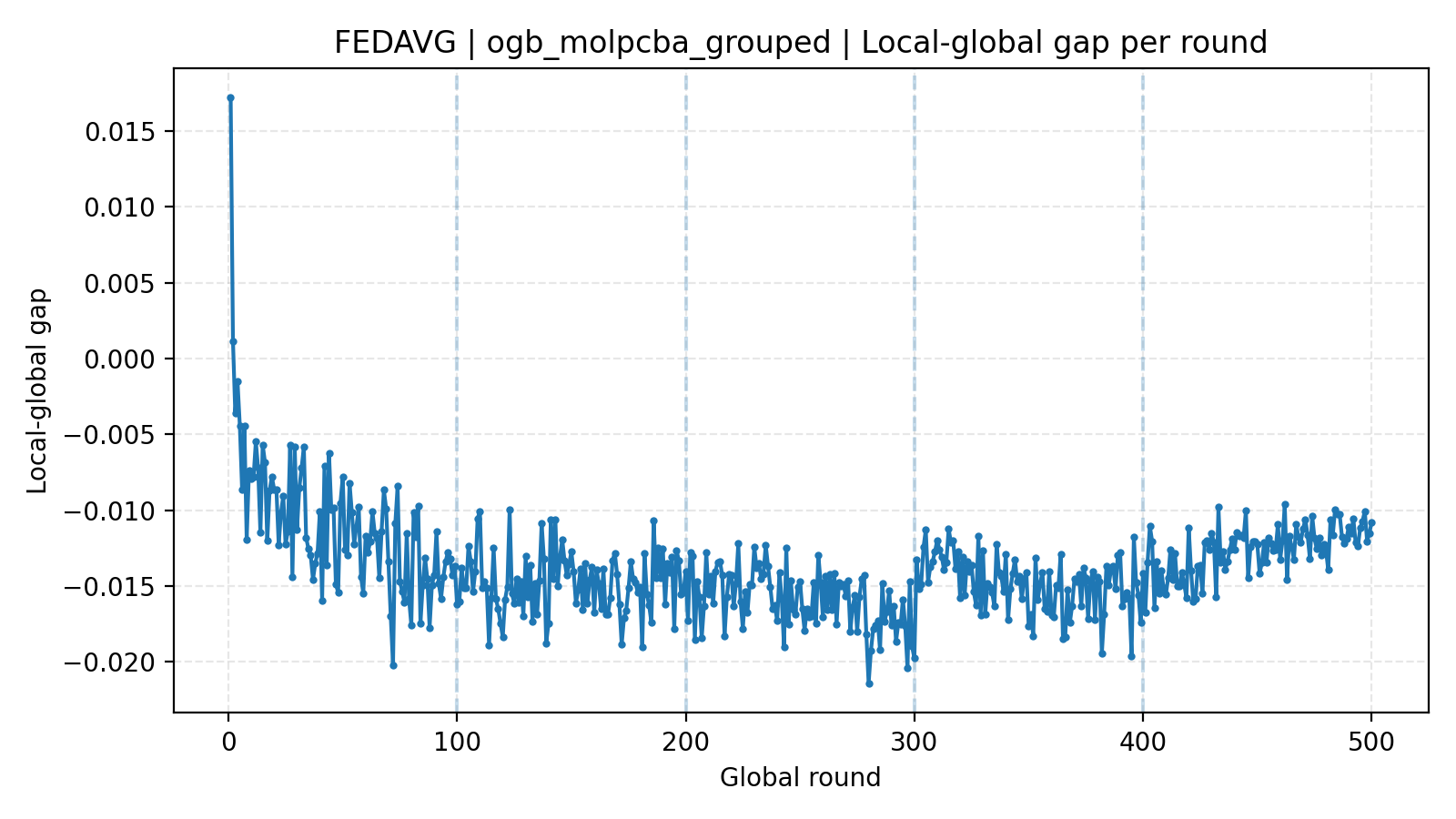}
        \caption{$K=5$ domains}
    \end{subfigure}
    \hfill
    \begin{subfigure}[t]{0.32\textwidth}
        \centering
        \includegraphics[width=\linewidth]{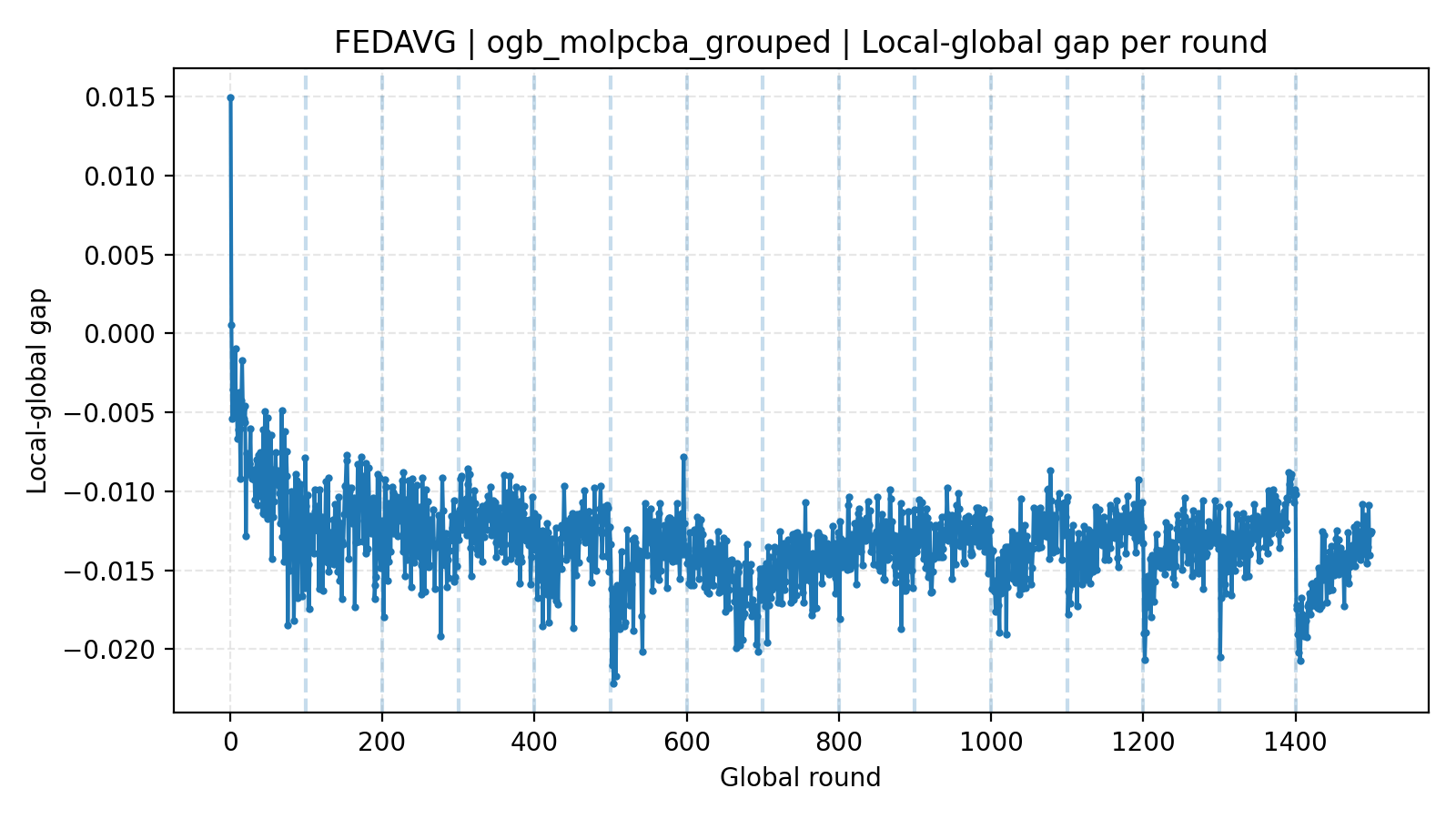}
        \caption{$K=15$ domains}
    \end{subfigure}
    \hfill
    \begin{subfigure}[t]{0.32\textwidth}
        \centering
        \includegraphics[width=\linewidth]{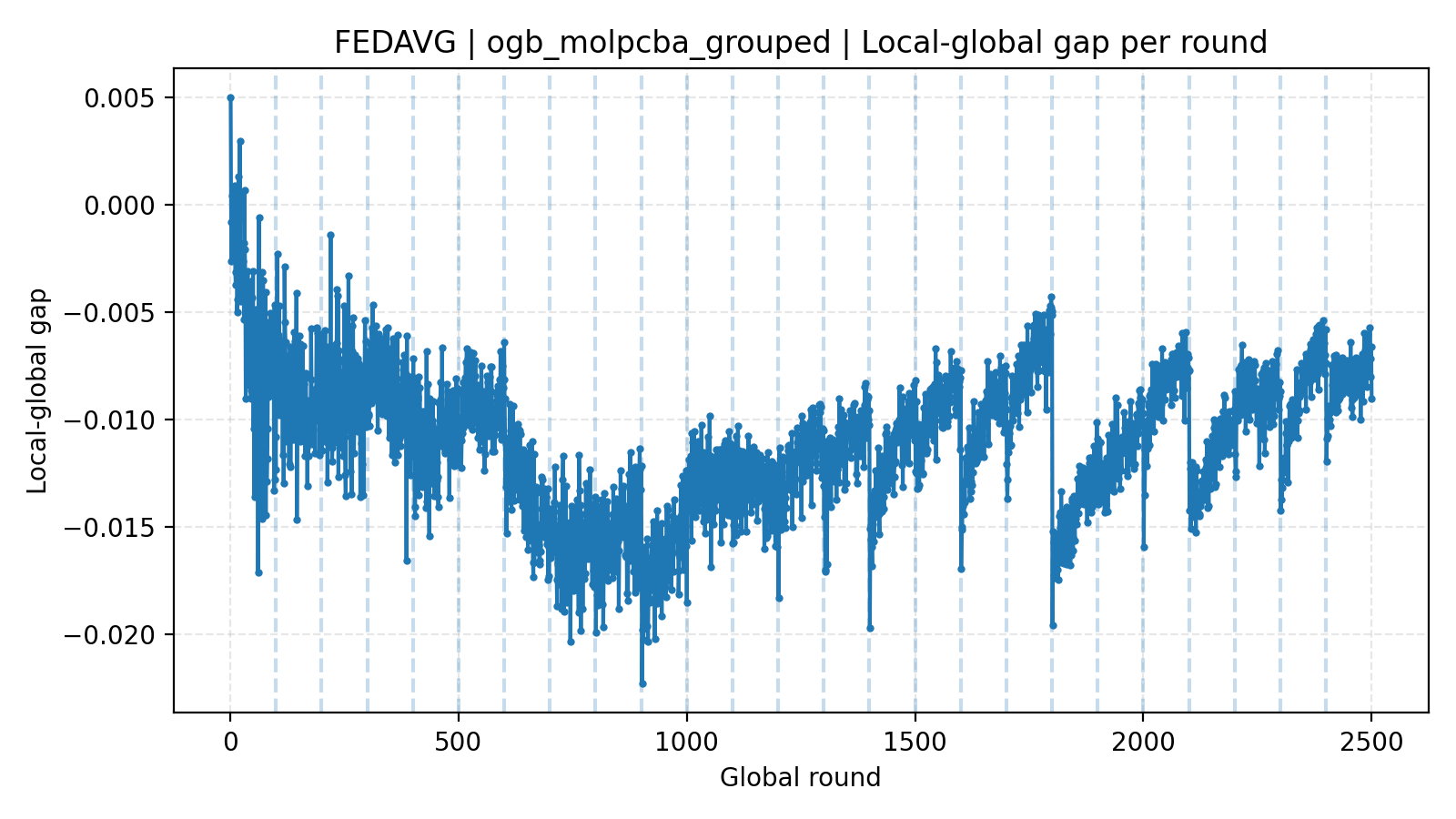}
        \caption{$K=25$ domains}
    \end{subfigure}
    \caption{Local-global gap on scaffold-grouped OGB-MolPCBA. The gap becomes more variable as domain granularity increases.}
    \label{fig:molpcba-gap}
\end{figure}

\section{Additional Ablations}
\label{app:additional_ablations}

This appendix includes non-core ablations that are useful for understanding implementation choices. These ablations are not part of the HERO-Core leaderboard.

\subsection{Homogeneous versus heterogeneous streams}

\begin{figure}[t]
\centering
\includegraphics[width=0.72\linewidth]{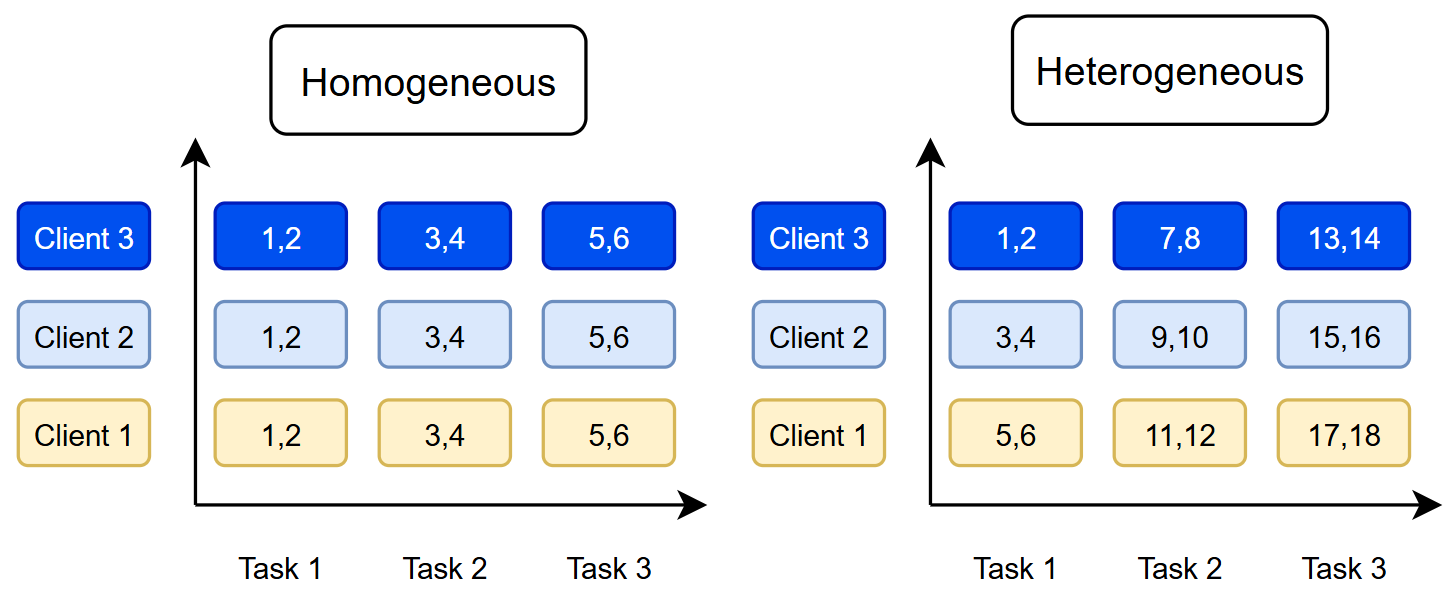}
\caption{Illustration of homogeneous and heterogeneous federated continual learning streams. In the homogeneous case, clients follow aligned class distributions. In the heterogeneous case, clients observe different class distributions or task sequences.}
\label{fig:homo-heter-overview}
\end{figure}

Figure~\ref{fig:homo-heter-overview} illustrates the distinction between homogeneous and heterogeneous streams. In homogeneous streams, clients have similar local objectives, so server aggregation receives more aligned updates. In heterogeneous streams, clients may observe different class distributions or task sequences, which makes local updates less aligned and can increase interference.

\begin{figure}[t]
    \centering
    \begin{subfigure}{0.48\textwidth}
        \centering
        \includegraphics[width=\linewidth]{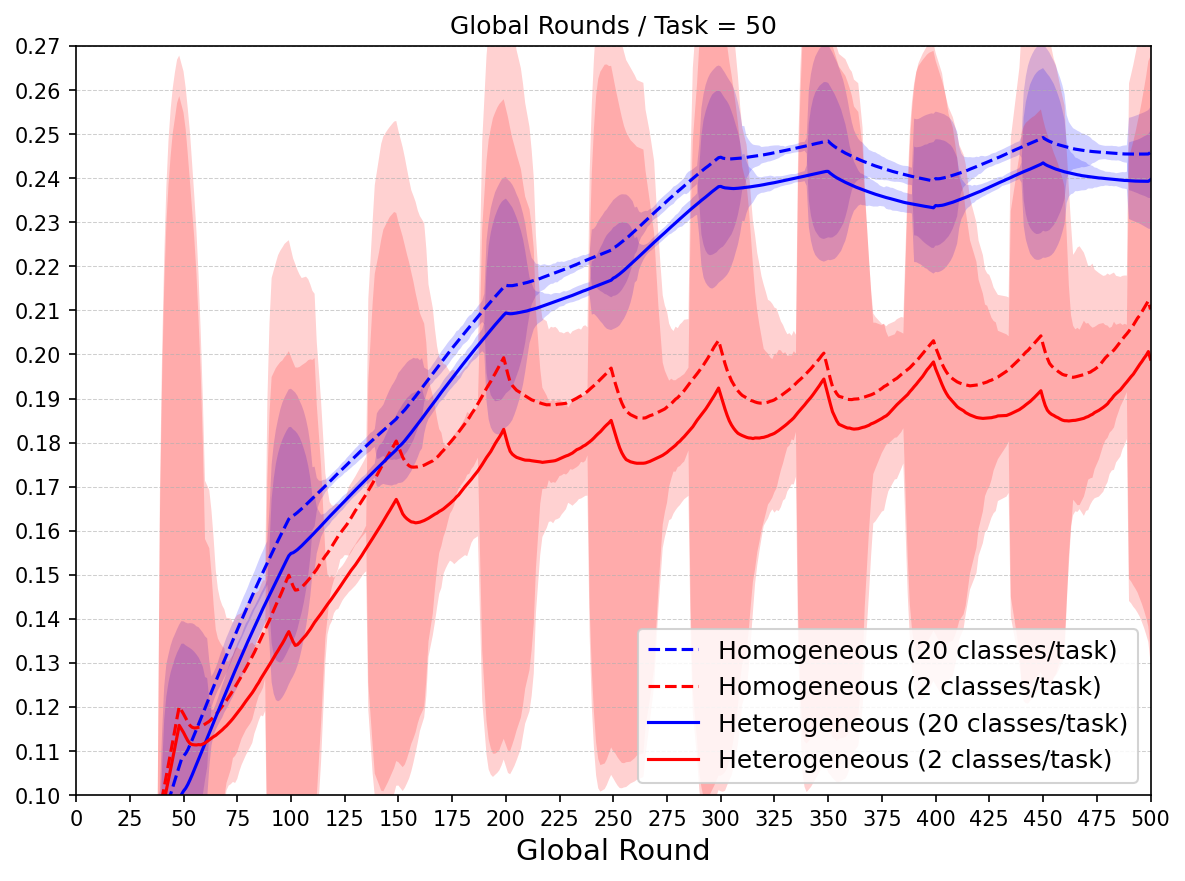}
        \caption{$50$ rounds per task}
    \end{subfigure}
    \hfill
    \begin{subfigure}{0.48\textwidth}
        \centering
        \includegraphics[width=\linewidth]{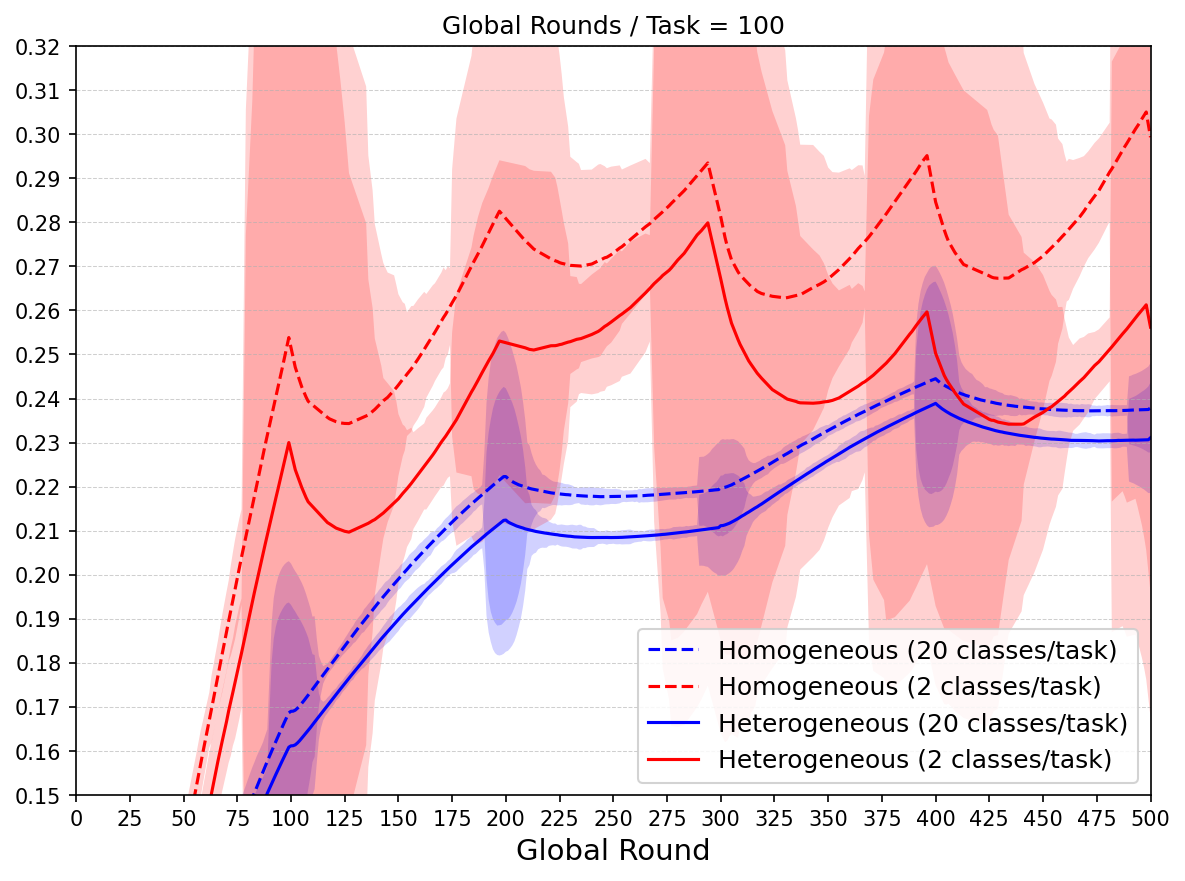}
        \caption{$100$ rounds per task}
    \end{subfigure}

    \vspace{0.5em}

    \begin{subfigure}{0.48\textwidth}
        \centering
        \includegraphics[width=\linewidth]{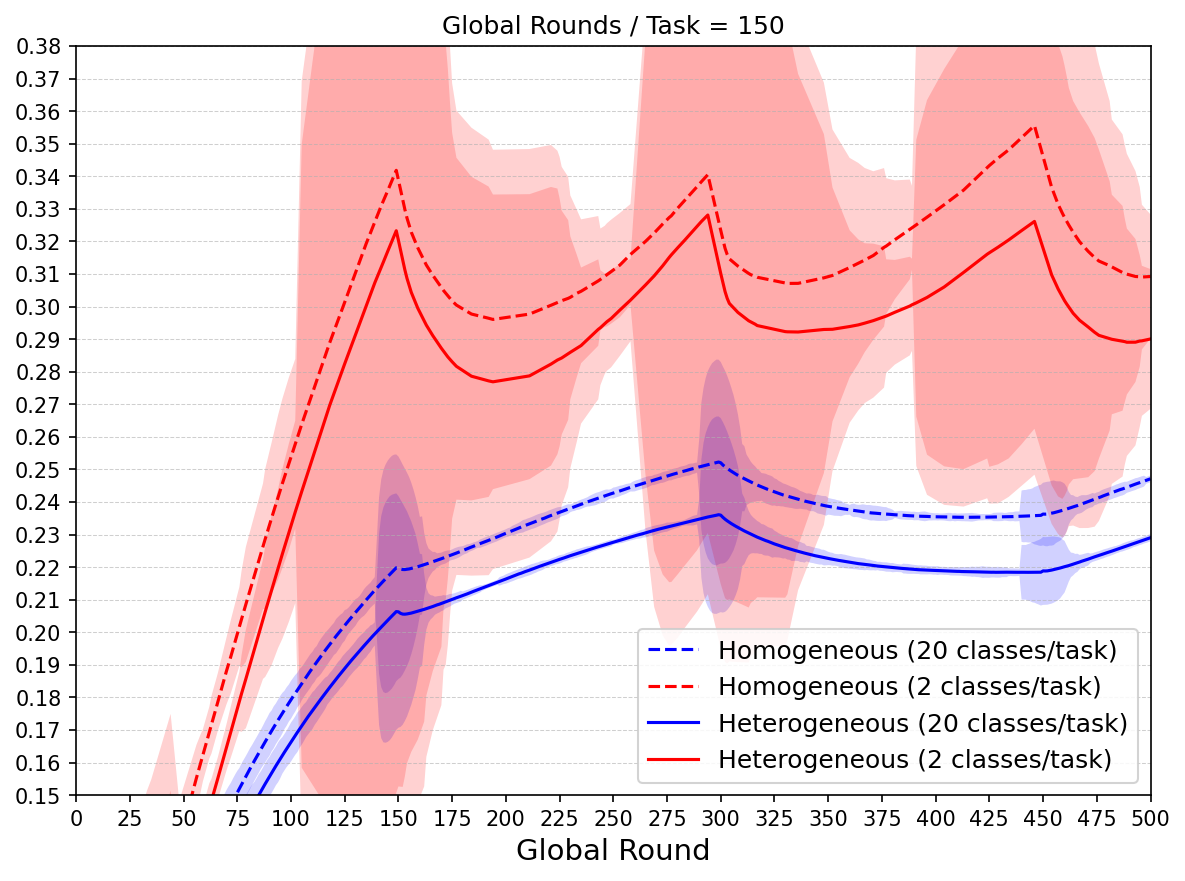}
        \caption{$150$ rounds per task}
    \end{subfigure}
    \hfill
    \begin{subfigure}{0.48\textwidth}
        \centering
        \includegraphics[width=\linewidth]{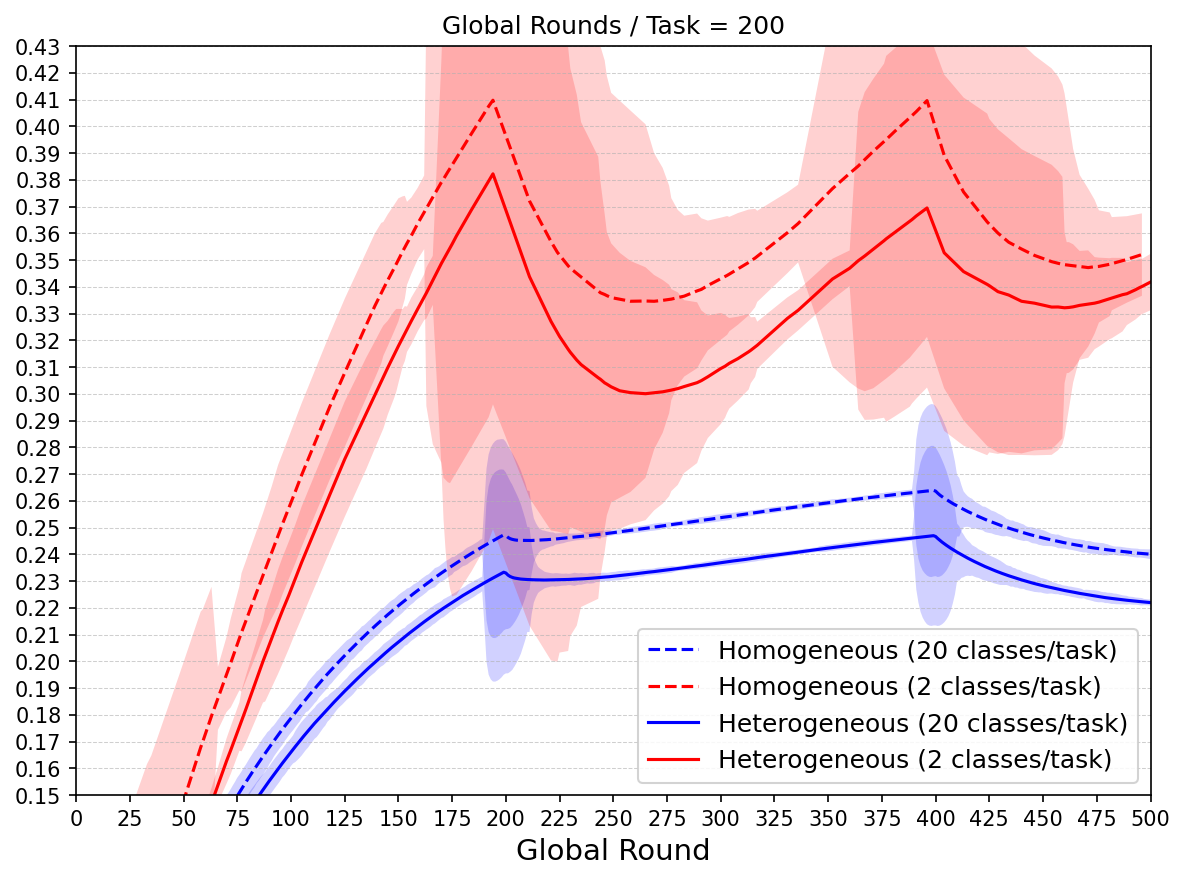}
        \caption{$200$ rounds per task}
    \end{subfigure}

    \caption{Homogeneous and heterogeneous stream comparisons under different communication budgets.}
    \label{fig:heter-homo}
\end{figure}

Figure~\ref{fig:heter-homo} compares homogeneous and heterogeneous streams under different numbers of global rounds per task. The homogeneous setting is more stable because client updates are more aligned. The heterogeneous setting shows larger fluctuations, especially when the communication budget increases and local training has more opportunity to specialize to biased client distributions. This ablation supports the use of controlled heterogeneity axes in HERO-Core.

\subsection{Full classifiers versus incremental classifiers}

\begin{figure}[t]
\centering
\includegraphics[width=0.70\linewidth]{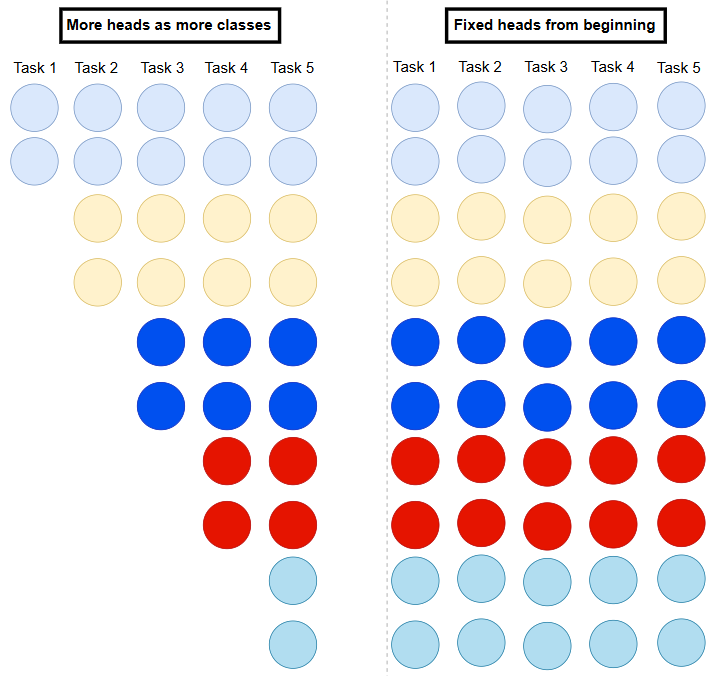}
\caption{Illustration of incremental classifiers and full classifiers. Incremental classifiers add heads as new classes arrive, while full classifiers include all heads from the beginning.}
\label{fig:incremental-heads}
\end{figure}

\begin{figure}[t]
    \centering
    \begin{subfigure}{0.48\textwidth}
        \centering
        \includegraphics[width=\linewidth]{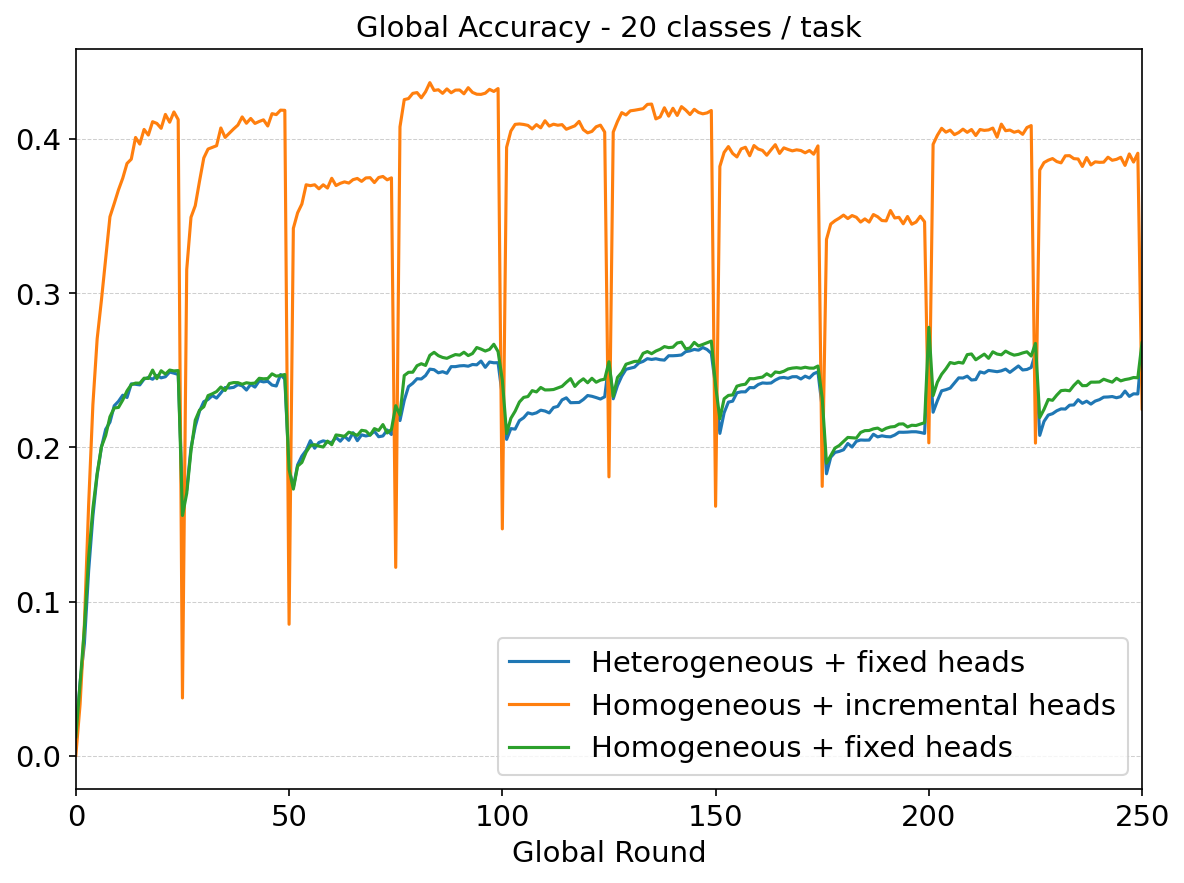}
        \caption{Global accuracy, $20$ classes/task}
    \end{subfigure}
    \hfill
    \begin{subfigure}{0.48\textwidth}
        \centering
        \includegraphics[width=\linewidth]{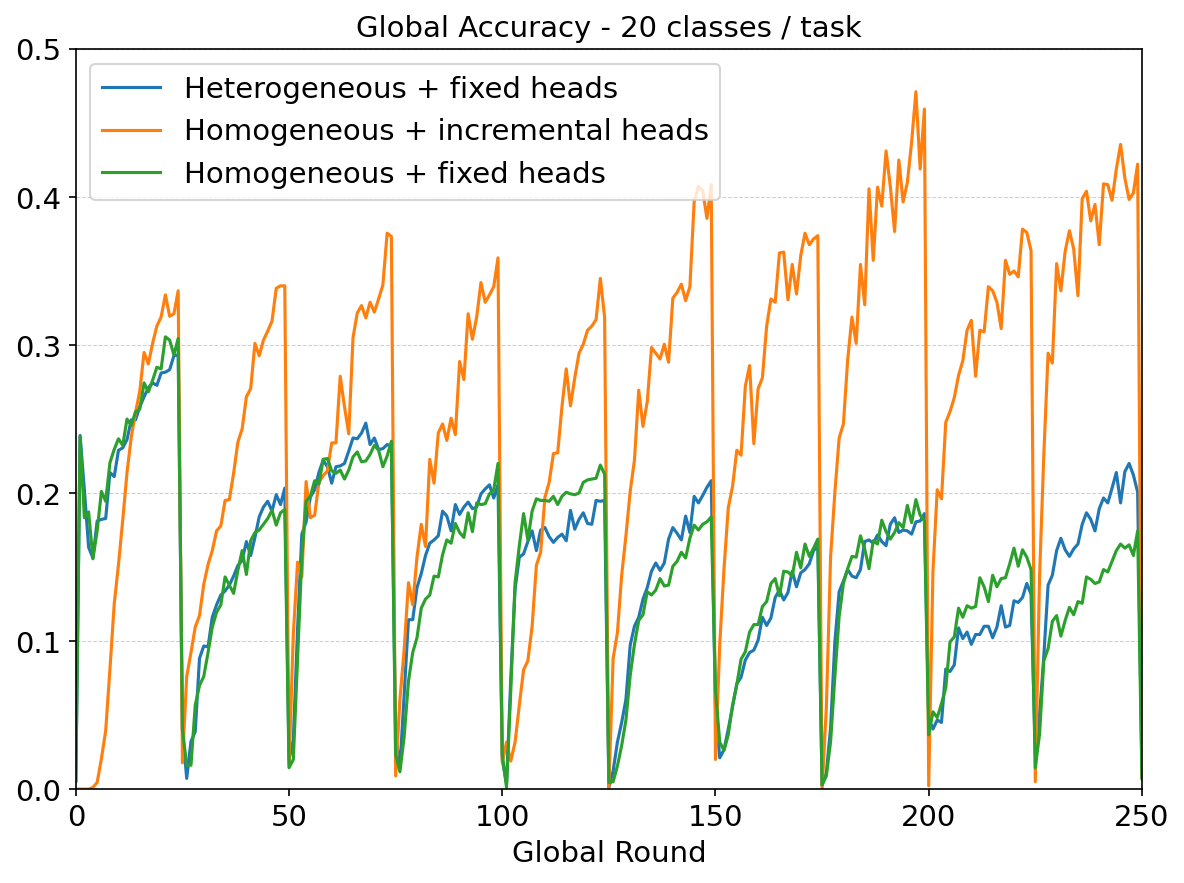}
        \caption{Global accuracy, $2$ classes/task}
    \end{subfigure}

    \vspace{0.5em}

    \begin{subfigure}{0.48\textwidth}
        \centering
        \includegraphics[width=\linewidth]{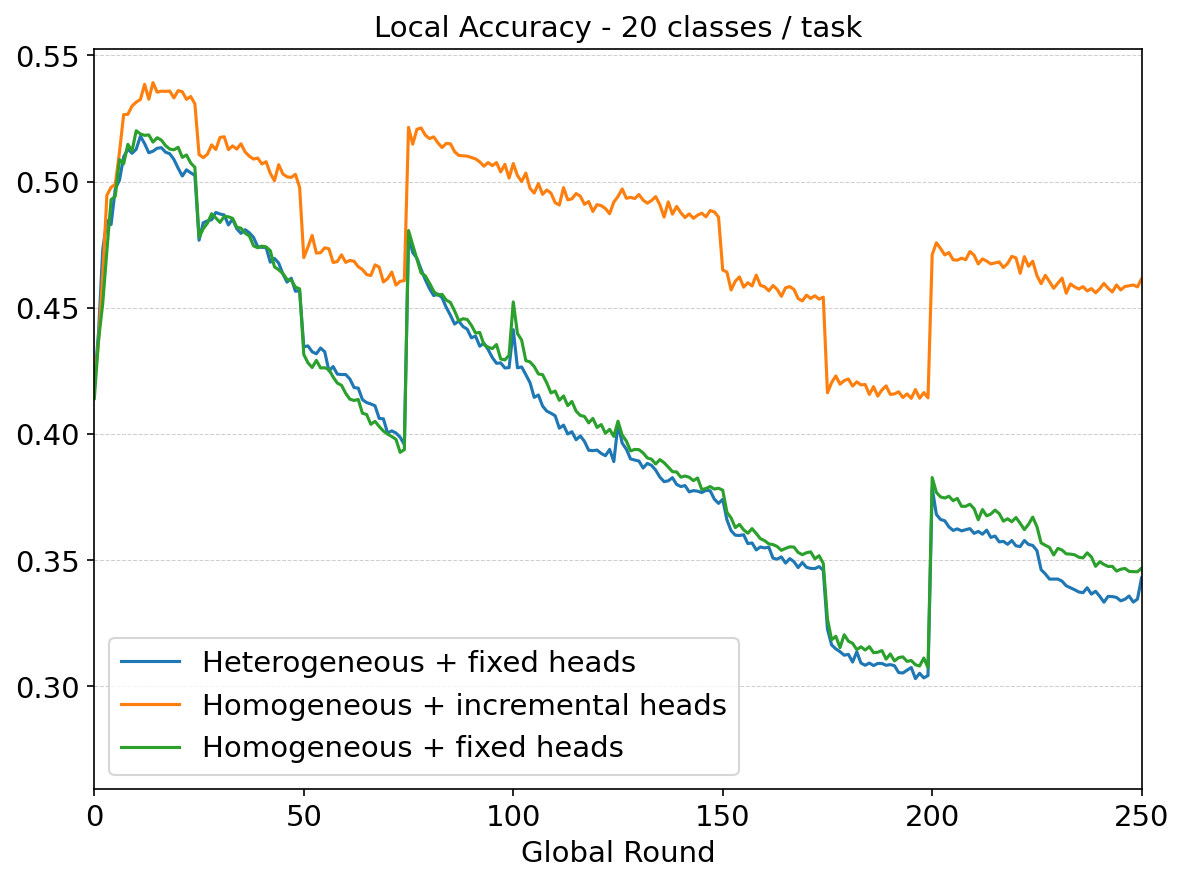}
        \caption{Local accuracy, $20$ classes/task}
    \end{subfigure}
    \hfill
    \begin{subfigure}{0.48\textwidth}
        \centering
        \includegraphics[width=\linewidth]{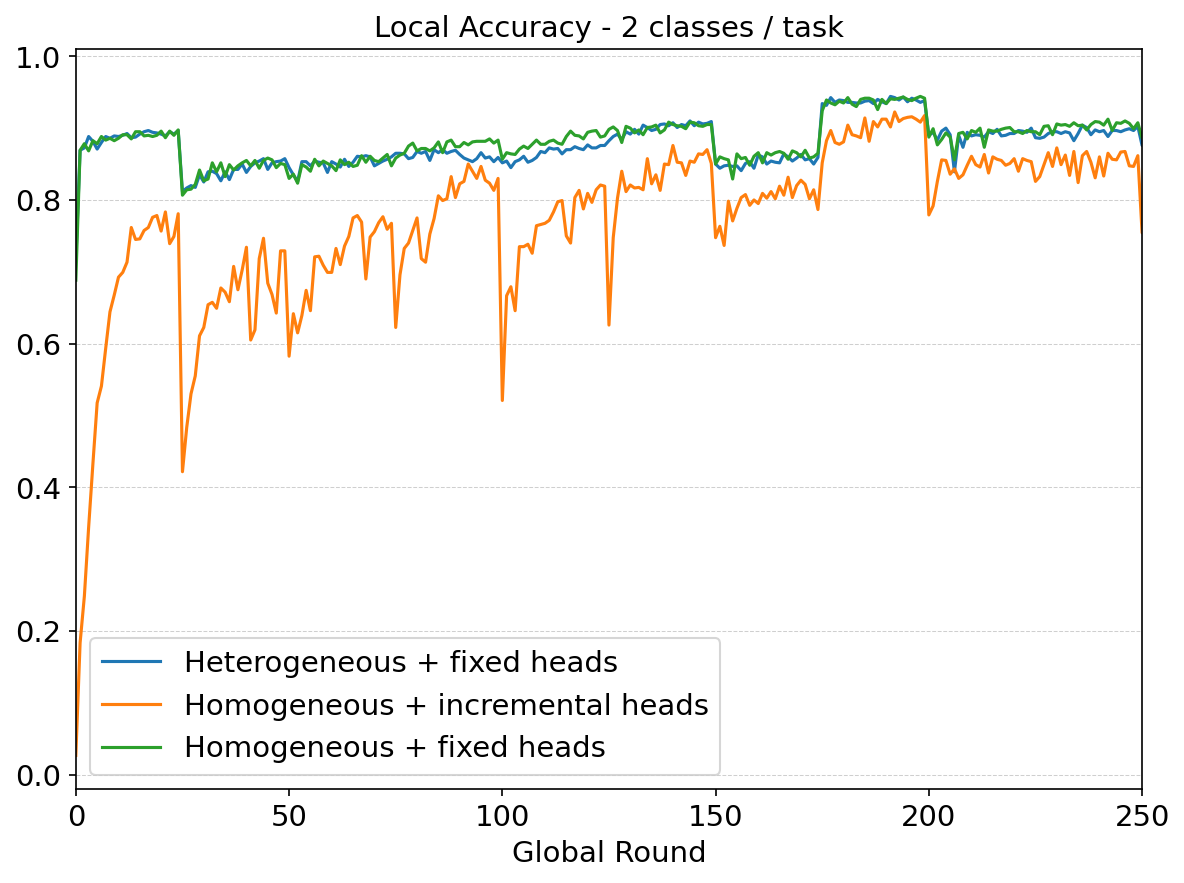}
        \caption{Local accuracy, $2$ classes/task}
    \end{subfigure}

    \caption{Comparison between full classifiers and incremental classifiers under global and local evaluation.}
    \label{fig:incre-full-global-local}
\end{figure}

Figure~\ref{fig:incremental-heads} illustrates the two classifier designs. In the full classifier setting, all class heads exist from the beginning, including heads for unseen classes. In the incremental classifier setting, new heads are introduced as new classes arrive. Figure~\ref{fig:incre-full-global-local} shows that incremental classifiers can improve global performance because the output space grows together with the class-incremental stream. Full classifiers may produce strong local accuracy in simple two-class tasks, but this does not necessarily translate to stronger global performance. This ablation motivates reporting global and client-aware metrics rather than relying only on local accuracy.

\end{document}